%% file: bayesiandip_arxiv.tex
\documentclass[10pt,twocolumn,letterpaper]{article}

\usepackage{cvpr}
\usepackage{times}
\usepackage{epsfig}
\usepackage{graphicx}
\usepackage{amsmath}
\usepackage{amsthm}
\usepackage{amssymb}
\usepackage{amsfonts}       % blackboard math symbols
\usepackage{bm}
\usepackage{placeins}

\usepackage[usenames]{color}
\newcommand{\ds}[1]{\normalmarginpar\marginpar{\scriptsize\color{red}{DS: #1 }}} % by default on right
\newcommand{\dsin}[1]{{\scriptsize \color{red}{[[DS: #1 ]]}}}
\newcommand{\sm}[1]{{\scriptsize \color{blue}{[[SM: #1 ]]}}} % by default on right
\newcommand{\new}[1]{#1} % by default on right
\newcommand{\newaxv}[1]{#1} % by default on right
\newcommand{\eat}[1]{}

\newcommand\blfootnote[1]{%
  \begingroup
  \renewcommand\thefootnote{}\footnote{#1}%
  \addtocounter{footnote}{-1}%
  \endgroup
}

\newcommand{\mypar}[1]{\vspace{3pt}\noindent{}\textbf{#1}}

\newcommand{\R}{\mathbb{R}}

\usepackage{booktabs}

\usepackage[pagebackref=true,breaklinks=true,letterpaper=true,colorlinks,bookmarks=false]{hyperref}

\cvprfinalcopy % *** Uncomment this line for the final submission

\def\cvprPaperID{3060} % *** Enter the CVPR Paper ID here

\ifcvprfinal\fi
\begin{document}

%%%%%%%%% TITLE
\title{A Bayesian Perspective on the Deep Image Prior}

\author{Zezhou Cheng \quad Matheus Gadelha \quad Subhransu Maji \quad
  Daniel Sheldon\\
University of Massachusetts, Amherst\\
{\tt\small \{zezhoucheng, mgadelha, smaji, sheldon\}@cs.umass.edu}
}

\maketitle
%\thispagestyle{empty}

%%%%%%%%% ABSTRACT
\begin{abstract}
The deep image prior~\cite{ulyanov18deep} was recently introduced as a prior for natural images. It represents images as the output of a convolutional network with random inputs. For ``inference'', gradient descent is performed to adjust network parameters to make the output match observations. This approach yields good performance on a range of image reconstruction tasks.
We show that the deep image prior is asymptotically equivalent to a stationary Gaussian process prior in the limit as the number of channels in each layer of the network goes to infinity, and derive the corresponding kernel. This informs a Bayesian approach to inference.
We show that by conducting posterior inference using stochastic gradient Langevin dynamics we avoid the need for early stopping, which is a drawback of the current approach, and improve results for denoising and impainting tasks. 
We illustrate these intuitions on a number of 1D and 2D signal reconstruction tasks.
% \blfootnote{\new{Code and supplementary materials are available at} \url{https://people.cs.umass.edu/~zezhoucheng/gp-dip}}
\blfootnote{\newaxv{Source code is available at} \url{https://people.cs.umass.edu/~zezhoucheng/gp-dip}}
\end{abstract}

%%%%%%%%% BODY TEXT
\input{intro}
\input{related}
\input{dipgp}

\input{experiments}

\input{conclusion}
\input{acknowledgement}
\vfill
\FloatBarrier
\clearpage
\input{appendix}

%\vfill
%\FloatBarrier
%\clearpage
\small{
\bibliography{bayesiandip}
\bibliographystyle{ieee_fullname}
}

% \newpage
\end{document}

%% file: intro.tex
\section{Introduction}
\label{sec:intro}
It is well known that deep convolutional networks trained on large datasets provide a rich hierarchical representation of images. Surprisingly, several works have shown that convolutional networks with \emph{random} parameters can also encode non-trivial image properties. For example, second-order statistics of filter responses of random convolutional networks are effective for style transfer and synthesis tasks~\cite{ustyuzhaninov2016texture}. On small datasets, features extracted from random convolutional networks can work just as well as trained networks~\cite{Saxe2011OnRW}. Along these lines, the \emph{``deep image prior"} proposed by Ulyanov \etal~\cite{ulyanov18deep} showed that the output of a suitably designed convolutional network on random inputs tends to be smooth and induces a natural image prior,
%The parameters of the network and its input parameterize the space of natural images, 
so that the search over natural images can be replaced by gradient descent to find network parameters and inputs to minimize a reconstruction error of the network output. Remarkably, no prior training is needed and the method operates by initializing the parameters randomly.

%given a noisy target $\bm{y}$ can be replaced by a search over the parameters $\bm{\theta} \in \mathbb{R}^D$ of a convolutional network $f$ with a fixed input $\bm{z}$ drawn from a distribution ${\cal D}_z$ as: $
%	\min_{\bm{\theta} \in \mathbb{R}^D} E\left(\bm{y}, f_{\bm{\theta}}(\bm{z})\right)$.
%The resulting optimization problem can be solved using gradient descent starting from a random initialization of the parameters. 
%In other words the network acts as the prior and the parameters index the space of natural images.

Our work provides a novel Bayesian view of the deep image prior.
%that sheds new light on its impressive performance.
%, by first proving theoretical properties of the implied random process, and then using Bayesian inference to address some of the drawbacks of current estimation techniques. 
We prove that a convolutional network with random parameters operating on a stationary input, \eg, white noise, approaches a two-dimensional Gaussian process (GP) with a \emph{stationary kernel} in the limit as the number of channels in each layer goes to infinity (Theorem~\ref{th:gp}).
While prior work~\cite{neal1995bayesian,williams1996gaussian,matthews2018gaussian,borovykh2018gaussian,novak2018bayesian} has investigated the GP behavior of infinitely wide networks and convolutional networks, our work is the first to analyze the spatial covariance structure induced by a convolutional network on stationary inputs. 
We analytically derive the kernel as a function of the network architecture and input distribution by characterizing the effects of convolutions, non-linearities, up-sampling, down-sampling, and skip connections on the spatial covariance.
These insights could inform choices of network architecture for designing 1D or 2D priors.

\begin{figure*}
    \centering
    \setlength{\tabcolsep}{1pt}
    \begin{tabular}{cccc}
    \includegraphics[height=0.17\linewidth]{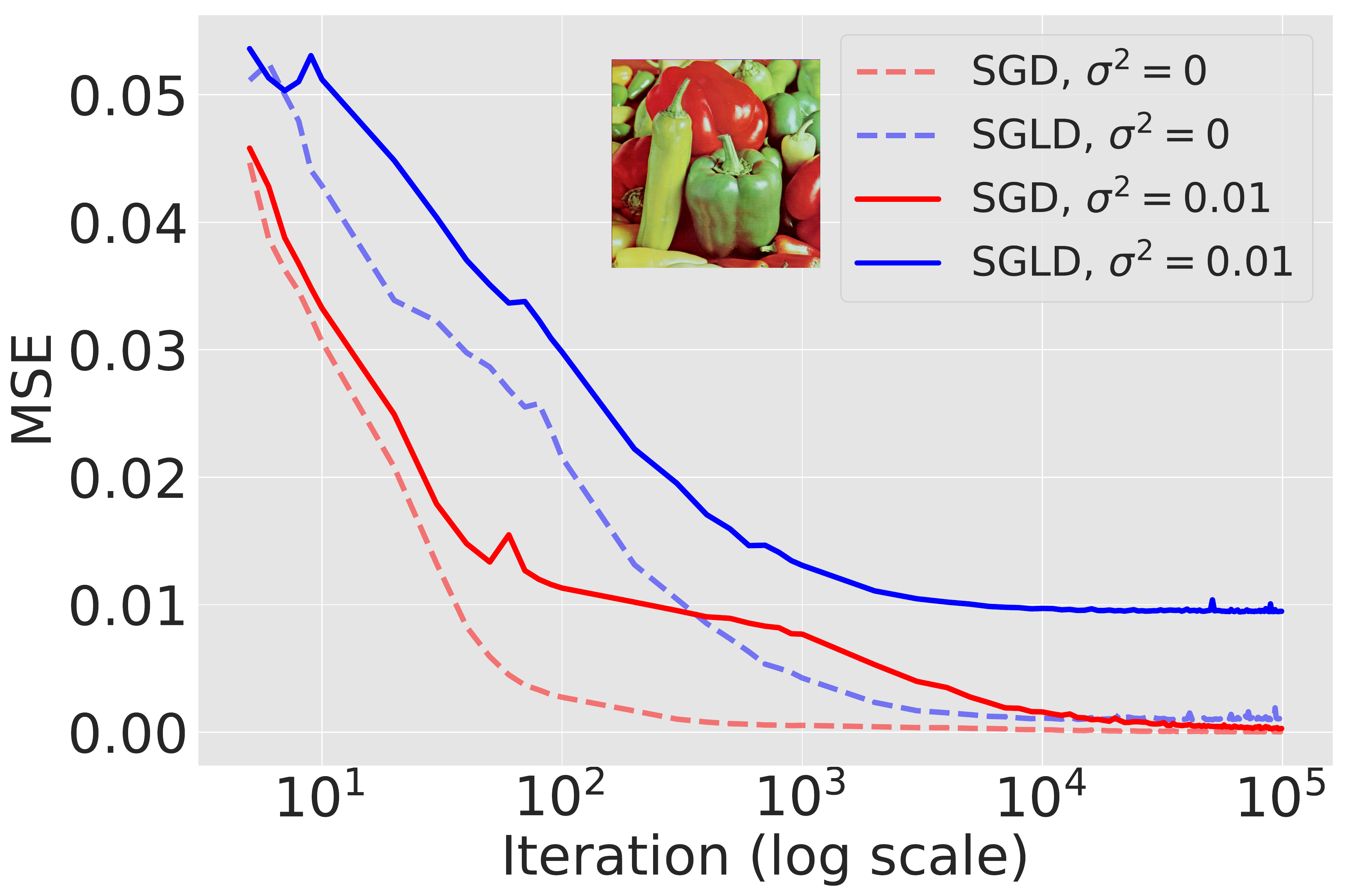} &
    \includegraphics[height=0.17\linewidth]{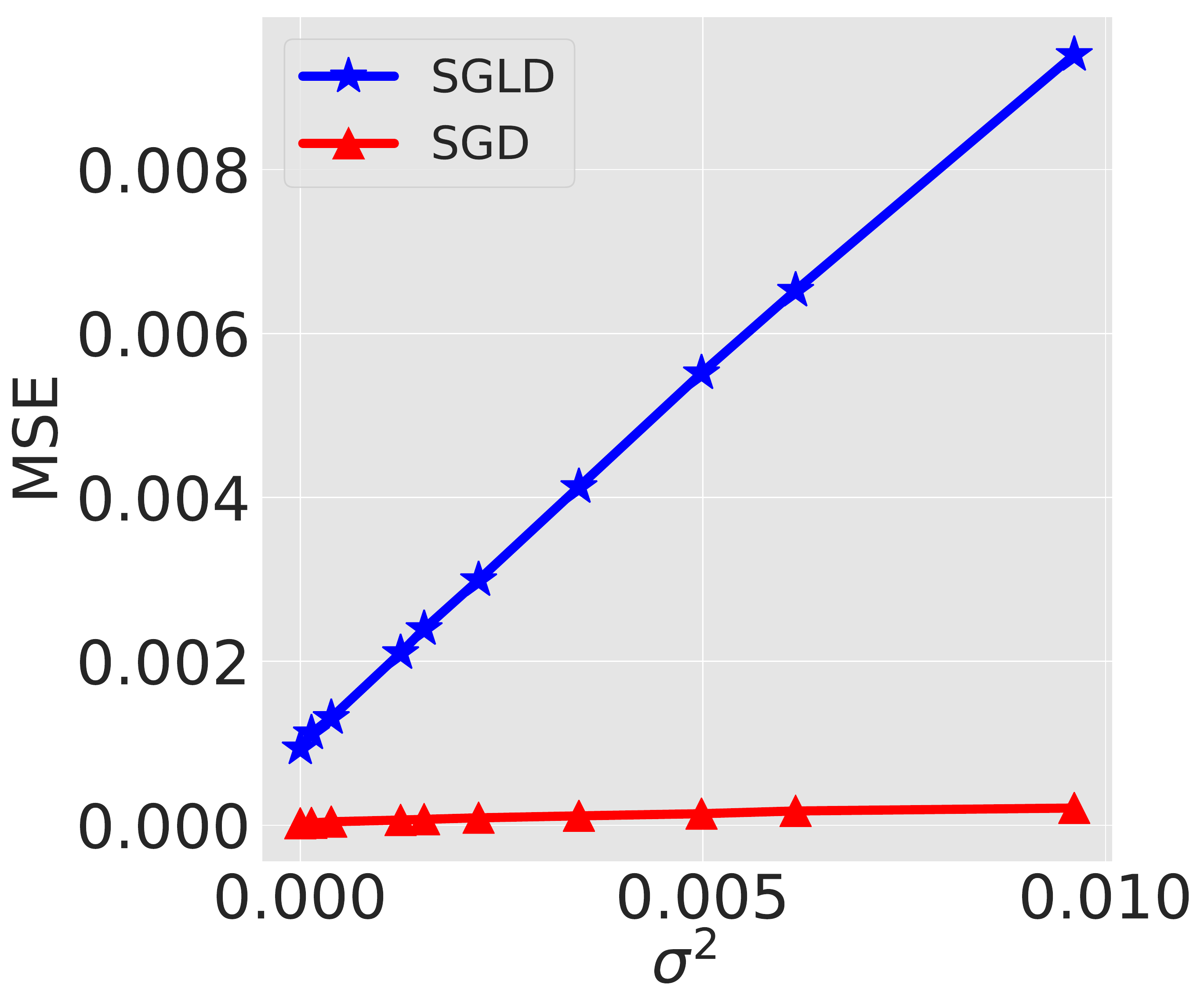} &
    \includegraphics[height=0.17\linewidth]{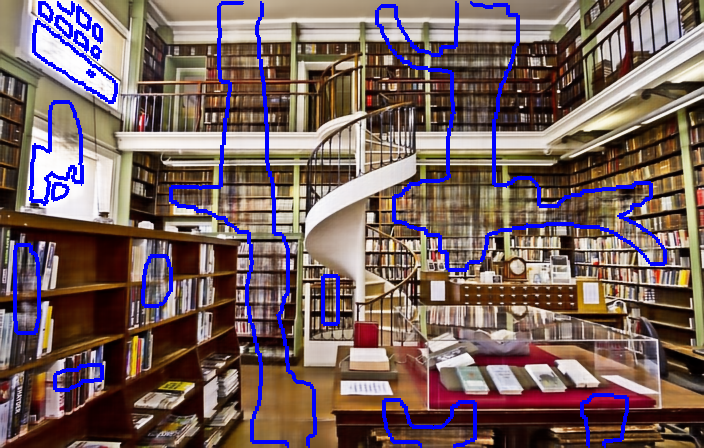} &
    \includegraphics[height=0.17\linewidth]{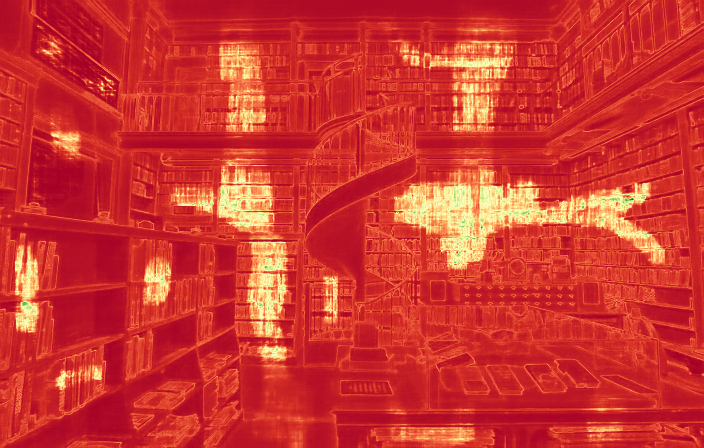} \\
    (a) MSE vs. Iteration & (b) Final MSE vs. $\sigma^2$ & (c) Inferred mean & (d) Inferred variance \\
    \end{tabular}
\caption{\emph{(Best viewed magnified.)} Denoising and inpainiting results with the deep image prior. \textbf{(a)} Mean Squared Error (MSE) of the inferred image with respect to the noisy input image as a function of iteration for two different noise levels. SGD converges to zero MSE resulting in overfitting while SGLD roughly converges to the noise level in the image. This is also illustrated in panel \textbf{(b)} where we plot the MSE of SGD and SGLD as a function of the noise level $\sigma^2$ after convergence. See Section~\ref{sec: denoising} for implementation details. \textbf{(c)} An inpainting result where parts of the image inside the blue boundaries are masked out and inferred using SGLD with the deep image prior. \textbf{(d)} An estimate of the variance obtained from posterior samples visualized as a heat map. Notice that the missing regions near the top left have lower variance as the area is uniform.}
% Inference with SGD to minimize the 

% Denoising experiments with the deep image prior. \textbf{(Left)} Peak Signal to Noise (PSNR) for denoising the "peppers" image using various inference schemes. SGD (blue curve) overfits, while sampling from the posterior using SGLD (green curve) and averaging (black) remains stable. 
% %The green curve is an intermediate variant where noise is only added to the input parameters which also overfits. 
% \textbf{(Right)} Mean Squared Error (MSE) for denoising four different images using SGD and SGLD. SGD eventually overfits every image while SGLD stabilizes at the error corresponding to the noise level of the input. \todo{This figure needs some work. Fix the color scheme so that we can sensibly refer to different methods. On the left figure we can perhaps get rid of the green (input noise version)}.}
% %PSNR curve of SGD variants and SGLD on denoising task. The target noisy image is generated by adding Gaussian noise with $\sigma = 25$ on the natural image (``peppers") presented in bottom right. Right: mean squared error (MSE) curve of SGD variants and SGLD on reconstruction task using: ``peppers", noisy ``peppers", randomly shuffled ``peppers" and white noise.}
\label{fig:splash-figure}
\vspace{-10pt}
\end{figure*}

We then use a Bayesian perspective to address drawbacks of current estimation techniques for the deep image prior.
Estimating parameters in a deep network from a single image poses a huge risk of overfitting.
In prior work the authors relied on early stopping to avoid this.
Bayesian inference provides a principled way to avoid overfitting 
by adding suitable priors over the parameters 
and then using posterior distributions to quantify uncertainty. 
However, posterior inference with deep networks is challenging.
\new{One option is to compute the posterior of the limiting GP.
For small networks with enough channels, we show this closely matches the deep image prior, but is computationally expensive.}
Instead, we conduct posterior sampling based on stochastic gradient Langevin dynamics (SGLD)~\cite{welling2011bayesian}, which is both theoretically well founded and computationally efficient, since it is based on standard gradient descent. 
We show that posterior sampling using SGLD avoids the need for early stopping and performs better than vanilla gradient descent on image denoising and inpainting tasks (see Figure~\ref{fig:splash-figure}).
It also allows us to systematically compute variances of estimates as a measure of uncertainty.
We illustrate these ideas on a number of 1D and 2D reconstruction tasks.

%% file: related.tex
\section{Related work}
\label{sec:related}
%We summarize related work on image priors, Gaussian processes, connections between neural networks and GPs, and Bayesian inference in  deep networks.
%\dsl{This section reads well and I'm leaving it for now. If time, it could potentially be rewritten as a historical narrative that weaves all of the threads together and provides more motivation and highlights significance of our work.}

%\paragraph{Image priors.}
\mypar{Image priors.}
Our work analyzes the \emph{deep image prior}~\cite{ulyanov18deep} that represents an image as a convolutional network $f$ with parameters $\theta$ on input $x$.
Given a noisy target $y$ the denoised image is obtained by minimizing the reconstruction error $\Vert y - f(x; \theta) \Vert$ over $x$ and $\theta$.
The approach starts from an initial value of $x$ and $\theta$ drawn i.i.d. from a zero mean Gaussian distribution and optimizes the objective through gradient descent, relying on early stopping to avoid overfitting (see Figure~\ref{fig:splash-figure}). 
Their approach showed that the prior is competitive with state-of-the-art learning-free approaches, such as BM3D~\cite{dabov2007image}, for image denoising, super resolution, and inpainting tasks.
The prior encodes hierarchical self-similarities that dictionary-based approaches~\cite{papyan2017convolutional} and non-local techniques such as BM3D and non-local means~\cite{buades2005non} exploit.
The architecture of the network plays a crucial role: several layer networks were used for inpainting tasks, while those with skip connections were used for denoising.
Our work shows that these networks induce priors that correspond to different smoothing ``scales".

%\paragraph{Gaussian processes (GPs).}
\mypar{Gaussian processes (GPs).}
A Gaussian processes is an infinite collection of random variables for which any finite subset are jointly Gaussian distributed~\cite{rasmussen2004gaussian}. A GP is commonly viewed as a prior over functions. Let $T$ be an index set (e.g., $T = \R$ or $T = \R^d$) and let $\mu(t)$ be a real-valued \emph{mean function} and $K(t, t')$ be a non-negative definite \emph{kernel} or \emph{covariance function} on $T$. If $f \sim GP(\mu, K)$, then, for any finite number of indices $t_1, \ldots, t_n \in T$, the vector $(f(t_i))_{i=1}^n$ is Gaussian distributed with mean vector $(\mu(t_i))_{i=1}^n$ and covariance matrix $(K(t_i, t_j))_{i,j=1}^n$. GPs have a long history in spatial statistics and geostatistics~\cite{matheron1973intrinsic}. In ML, interest in GPs was motivated by their connections to neural networks (see below). GPs can be used for general-purpose Bayesian regression~\cite{williams1996gaussian,rasmussen1999evaluation}, classification~\cite{williams1998bayesian}, and many other applications~\cite{rasmussen2004gaussian}.

%\paragraph{Deep networks and GPs.}
\mypar{Deep networks and GPs.}
Neal~\cite{neal1995bayesian} showed that a two-layer network converges to a Gaussian process as its width goes to infinity.
%and derived its covariance function. 
Williams~\cite{williams1997computing} provided expressions for the covariance function of networks with sigmoid and Gaussian transfer functions.
Cho and Saul~\cite{cho2009kernel} presented kernels for the ReLU and the Heaviside step non-linearities and investigated their effectiveness with kernel machines.
Recently, several works~\cite{lee2018deep,matthews2018gaussian} have extended these results to deep networks and derived covariance functions for the resulting GPs.
Similar analyses have also been applied to convolutional networks.
Garriaga-Alonso \etal\cite{garriga2018deep} investigated the GP
behavior of convolutional networks with residual layers, while
Borovykh~\cite{borovykh2018gaussian} analyzed the covariance functions
in the limit when the filter width goes to infinity. \new{Novak
  \etal\cite{novak2018bayesian} evaluated the effect of pooling layers
  in the resulting GP.}
Much of this work has been applied to prediction tasks, where given a dataset ${\cal D} = \{(x_i, y_i)\}_{i=1}^n$, a covariance function induced by a deep network is used to estimate the posterior $p(y|x,{\cal D})$ using standard GP machinery.
%\ds{I added ${\cal D}$ on the RHS. Is that what you meant?}
In contrast, we view a convolutional network as a spatial random process over the image coordinate space and study the induced covariance structure.
%over random spatial inputs as a family of functions indexed by its parameters and inputs and study the spatial covariance function induced by it.

%\paragraph{Bayesian inference with deep networks.}
\mypar{Bayesian inference with deep networks.}
It has long been recognized that Bayesian learning of neural networks
weights would be
desirable~\cite{mackay1992practical,neal1995bayesian}, \eg, to
prevent overfitting and quantify uncertainty. Indeed, this was a
motivation in the original work connecting neural networks and GPs. 
%This generally takes the form of placing a prior over the weights and then computing a posterior distribution given data. 
Performing MAP estimation with respect to a prior on the weights is
computationally straightforward and corresponds to
regularization. However, the computational challenges of full
posterior inference are significant. Early works used
MCMC~\cite{neal1995bayesian} or the Laplace
approximation~\cite{denker1991transforming,mackay1992practical} but
were much slower than basic learning by backpropagation. Several
variational inference (VI) approaches have been proposed over the
years~\cite{hinton1993keeping,Barber1998,graves2011practical,blundell2015weight}. Recently,
dropout was shown to be a form of approximate Bayesian
inference~\cite{gal2016dropout}. The approach we will use is based on
stochastic gradient Langevin dynamics
(SGLD)~\cite{welling2011bayesian}, a general-purpose method to convert
SGD into an MCMC sampler by adding noise to the iterates. Li
\etal~\cite{li2016preconditioned} describe a preconditioned SGLD method
for deep networks.

%% file: dipgp.tex
\section{Limiting GP for Convolutional Networks}
Previous work focused on the covariance of (scalar-valued) network outputs for two different inputs (\ie, images). For the deep image prior, we are interested in the \emph{spatial} covariance structure within each layer of a convolutional network. As a basic building block, we consider a multi-channel input image $X$ transformed through a convolutional layer, an elementwise non-linearity, and then a second convolution to yield a new multi-channel ``image'' $Z$, and derive the limiting distribution of a representative channel $z$ as the number of input channels and filters go to infinity.
First, we derive the limiting distribution when $X$ is \emph{fixed}, which mimics derivations from previous work. We then let $X$ be a stationary random process, and show how the spatial covariance structure propagates to $z$, which is our main result. We then apply this argument inductively to analyze multi-layer networks, and also analyze other network operations such as upsampling, downsampling, etc.

\setlength{\belowdisplayskip}{4.5pt} 
\setlength{\belowdisplayshortskip}{4.5pt}
\setlength{\abovedisplayskip}{4.5pt} 
\setlength{\abovedisplayshortskip}{4.5pt}

\subsection{Limiting Distribution for Fixed $X$}
\label{sec:fixed-x}
For simplicity, consider an image $X \in \mathbb{R}^{c\times T}$ with
$c$ channels and only one spatial dimension. The derivations are
essentially identical for two or more spatial dimensions. The first
layer of the network has $H$ filters denoted by $U = (u_1, u_2, \ldots
u_H)$ where $u_k \in \mathbb{R}^{c\times d}$ and the second layer has
one filter $v \in \mathbb{R}^H$ (corresponding to a single channel of
the output of this layer). The output of this network is:
\begin{align*}
  z = v * h(X*U) = \sum_{k=1}^{H} v_k h(X*u_k).
\end{align*}
The output $z = (z(1), z(2), \ldots, z(T'))$ also has one spatial dimension. Following~\cite{neal1995bayesian,williams1997computing} we derive the distribution of $z$ when $U \sim N(0, \sigma_u^2\mathbb{I})$ and $v \sim N(0, \sigma_v^2\mathbb{I})$. The mean is
\begin{equation*}
\begin{split}
\mathbb{E}[z(t)] &= \mathbb{E} \left[ \sum_{k=1}^{H} v_k h\left((X * u_k)(t)\right)\right]  \\
  &= \mathbb{E}\left[\sum_{k=1}^{H} v_k h\left(\sum_{i=1,j=1}^{c,d} x(i, t+1-j){u}_k(i,j)\right)
  \right] .
\end{split}
\end{equation*}
By linearity of expectation and independence of $u$ and $v$,
$$
\mathbb{E}[z(t)] = \sum_{k=1}^H \mathbb{E}[v_k] \mathbb{E}\left[ h\left((X * u_k)(t)\right) \right]  = 0, 
$$
since $v$ has a mean of zero. 
The central limit theorem (CLT) can be applied when $h$ is bounded to show that $z(t)$ approaches in distribution to a Gaussian as $H \rightarrow \infty$ and $\sigma^2_v$ is scaled as $1/H$.
Note that $u$ and $v$ don't need to be Gaussian for the CLT to apply, but we will use this property to derive the covariance. This is given by
\begin{equation*}
\begin{split}
K_z(t_1,t_2) = {} & \mathbb{E}[z(t_1)z(t_2)] \\
%= {} & \mathbb{E} \Biggr[ 
%                    \left(\sum_{k=1}^{H} v_k h\big((X * u_k)(t_1)\big)\right) \times \\
% {} &  \quad\quad \left(\sum_{k=1}^{H} v_k h\big((X * u_k)(t_2)\big)\right) 
%                  \Biggr] \\
 = {} & \mathbb{E} \left[ 
                        \sum_{k=1}^{H} v_k^2 h\big((X * u_k)(t_1)\big)
                                             h\big((X * u_k)(t_2)\big) 
                    \right] \\
= {} & H\sigma_v^2 \mathbb{E} \left[ 
                                h\big((X * u_1)(t_1)\big)
                                h\big((X * u_1)(t_2)\big)
                              \right].
\end{split}
\end{equation*}
The last two steps follow from the independence of $u$ and $v$ and that $v$ is drawn from a zero mean Gaussian.
Let $\bar{x}(t) = {\rm vec}\left( [X(:, t),X(:, t-1), \ldots, X(:, t-d+1)] \right)$ be the flattened tensor with elements within the window of size $d$ at position $t$ of $X$. 
%Here ${\rm vec}(\cdot)$ denotes the vector operator. 
Similarly denote $\bar{u} = {\rm vec}(u)$. 
Then the expectation can be written as
\begin{align}
  K_z(t_1, t_2) = H\sigma_v^2 E_u \left[ h(\bar{x}(t_1)^T\bar{u})h(\bar{x}(t_2)^T\bar{u}) \right].
\end{align}
Williams~\cite{williams1997computing} showed $V(x, y) =E_u \left[ h(x^Tu)h(y^Tu)\right]$ can be computed analytically for various transfer functions. 
For example, when $h(x) = {\rm erf}(x) = 2/\sqrt{\pi}\int_{0}^{x}e^{-s^2}ds$, then 
\begin{align}  \label{eq:2}
V_{{\rm erf}}(x, y) = \frac{2}{\pi}\sin^{-1} \frac{x^T\Sigma y}{\sqrt{\left(x^T\Sigma x \right)\left(y^T\Sigma y\right)}}. 
\end{align}
Here $\Sigma=\sigma^2 \mathbb{I}$ is the covariance of $u$.
Williams also derived kernels for the Gaussian transfer function $h(x, u) = \exp\{-(x-u)^T(x-u)/2\sigma^2\}$. 
For the ReLU non-linearity, \ie,  $h(t) = \max(0,t)$, Cho and Saul~\cite{cho2009kernel} derived the expectation as:
\begin{align} 
V_{\rm relu}(x, y) = \frac{1}{2\pi}\Vert x \Vert \Vert y \Vert \big(\sin \theta + (\pi -\theta)\cos \theta \big), 
\end{align}
where $\theta = \cos^{-1}\left(\frac{x^Ty}{\Vert x \Vert  \Vert y \Vert}\right)$. We refer the reader to~\cite{cho2009kernel,williams1997computing} for expressions corresponding to other transfer functions.

Thus, letting $\sigma_v^2$ scale as $1/H$ and $H \rightarrow \infty$ and for any input $X$, the output $z$ of our basic convolution-nonlinearity-convolution building block converges to a Gaussian distribution with zero mean and covariance 
\begin{equation} \label{eq:var_scale}
  K_z(t_1, t_2) = V\left(\bar{x}(t_1), \bar{x}(t_2)\right).
\end{equation}
%
%The covariance function depends on the properties of the input $X$ and the type of non-linearity.

\subsection{Limiting Distribution for Stationary $X$}
\label{sec:stationary}
%Now lets derive the mean and covariance function of $Z$ when 
We now consider the case when channels of $X$ are drawn i.i.d. from a stationary distribution. A signal $x$ is stationary (in the weak- or wide-sense) if the mean is position invariant and the covariance is shift invariant, \ie,
\begin{align}
m_x = \mathbb{E}[x(t)] = \mathbb{E}[x(t+\tau)],~~\forall \tau
\end{align}
and
\begin{equation}
\begin{split}
K_x(t_1,t_2) &= \mathbb{E}[(x(t_1) - m_x)(x(t_2) - m_x)] \\
             &= K_x(t_1 - t_2), \forall t_1, t_2.
\end{split}
\end{equation}
An example of a stationary distribution is white noise where $x(i)$ is
i.i.d. from a zero mean Gaussian distribution $N(0, \sigma^2)$
resulting in a mean $m_x = 0$ and covariance $K_x(t_1, t_2) = \sigma^2
\mathbf{1}[t_1=t_2]$.
Note that the input for the deep image prior is drawn from this distribution.

\newtheorem{theorem}{Theorem}
\begin{theorem}
\label{th:gp}
Let each channel of $X$ be drawn independently from a zero mean stationary distribution with covariance function $K_x$. Then the output of a two-layer convolutional network with the sigmoid non-linearity, \ie, $h(t) = {\rm erf}(t)$, converges to a zero mean stationary Gaussian process as the number of input channels $c$ and filters $H$ go to infinity \new{sequentially}. The stationary covariance $K_z$ is given by
$$
K_z^{\rm erf} (t_1, t_2) = K_z(r) = \frac{2}{\pi}\sin^{-1} \frac{K_x(r)}{K_x(0)}.
$$
where $r = t_2 - t_1$. 
\end{theorem}  \newcommand{\E}{\mathbb{E}}
\new{The full proof is included in the \newaxv{Appendix} and is obtained
  by applying the continious mapping thorem~\cite{mann1943stochastic}
  on the formula for the sigmoid non-linearity. The theorem implies that the limiting distribution
  of $Z$ is a stationary GP if the input $X$ is stationary.}

\newtheorem{lemma}{Lemma}
\begin{lemma} 
\label{lemma:relu}
Assume the same conditions as Theorem~\ref{th:gp} except the non-linearity is replaced by ReLU. Then the output converges to a zero mean stationary Gaussian process with covariance $K_z$
\begin{equation} \label{eq:relu-recurrence}
K_z^{\rm relu}(t_1, t_2) = \frac{K_x(0)}{2\pi}\Big( \sin\theta^x_{t_1,t_2} + (\pi - \theta^x_{t_1,t_2})\cos \theta^x_{t_1,t_2}\Big),
\end{equation}
where $\theta^x_{t_1,t_2} = \cos^{-1}\left(K_x(t_1,t_2)/K_x(0)\right)$. In terms of the angles we get the following:
\begin{align*}
\cos\theta^z_{t_1,t_2} = \frac{1}{\pi}\Big( \sin\theta^x_{t_1,t_2} + (\pi - \theta^x_{t_1,t_2})\cos \theta^x_{t_1,t_2} \Big).
\end{align*}
\end{lemma}

This can be proved by applying the recursive formula for ReLU non-linearity~\cite{cho2009kernel}. 
One interesting observation is that, for both non-linearities, the output covariance $K_z(r)$ at a given offset $r$ only depends on the input covariance $K_x(r)$ at the same offset, and on $K_x(0)$. 

\mypar{Two or more dimensions.} The results of this section hold without modification and essentially the same proofs for inputs with $c$ channels and two or more spatial dimensions by letting $t_1$, $t_2$, and $r = t_2 - t_1$ be vectors of indices.

\subsection{Beyond Two Layers}
So far we have shown that the output of our basic two-layer building block converges to a zero mean stationary Gaussian process as \new{$c \to \infty$ and then $H \to \infty$}. Below we discuss the effect of adding more layers to the network.

\mypar{Convolutional layers.} A proof of GP convergence for deep
networks was presented in~\cite{matthews2018gaussian}, including the
case for transfer functions that can be bounded by a \emph{linear
  envelope}, such as ReLU. In the convolutional setting, this implies
that the output converges to GP as the number of filters
in each layer simultaneously goes to infinity.
The covariance function can be obtained by recursively applying Theorem~\ref{th:gp} and Lemma~\ref{lemma:relu}; stationarity is preserved at each layer.
%Since each recursion leads to a stationary covariance the output of the final layer is also stationary.

\mypar{Bias term.} Our analysis holds when a bias term $b$ sampled from a zero-mean Gaussian is added, \ie, $z^{\rm bias} = z + b$. In this case the GP is still zero-mean but the covariance function becomes: $K^{\rm bias}_z(t_1, t_2) = \sigma_b^2  + K_z(t_1, t_2) $, which is still stationary.
%\ds{I don't understand the part "assume the signals we aim to generate are band-limited"}

\mypar{Upsampling and downsampling layers.} 
Convolutional networks have upsampling and downsampling layers to induce hierarchical representations. 
It is easy to see that downsampling (decimating) the signal preserves
stationarity since $K^{\downarrow}_x(t_1, t_2) = K_x(\tau t_1, \tau
t_2)$ where $\tau$ is the downsampling factor. 
Downsampling by average pooling also preserves stationarity. The
resulting kernel can be obtained by applying a uniform filter corresponding
to the size of the pooling window, which results in a stationary
signal, followed by downsampling.
%\new{Downsampling by
%  average pooling preserves stationarity since the uniform filter is
%  stationary.}\ds{I'm not sure what this means.}
However, upsampling in general does not preserve stationarity.
%For example, upsampling using nearest-neighbor interpolation makes the signal cyclostationary with a period equal to the upsampling factor. 
Therrien~\cite{therrien2001issues} describes the conditions under which upsampling a signal with a linear filter maintains stationarity. In particular, the upsampling filter must be band limited, such as the ${\rm sinc}$ filter: ${\rm sinc}(x) = \sin(x)/x$.
%However band-limited filters cannot be spatially limited are therefore computationally inefficient.
If stationarity is preserved the covariance in the next layer is given
by $K^{\uparrow}_x(t_1, t_2) = K_x(t_1/\tau, t_2/\tau)$.

\eat{ 
\ds{I'm
  slightly confused by ``if stationarity is preserved'' or what the
  ``upsampling filter'' is here. I think there may be a disconnect
  between the operation of upsampling which must happen on grid-based
  samples, and the mathematical operations of upsampling which just
  divides the lags by $\tau$.}
\sm{Here is my understanding: For upsampling a signal we can first
  put the signal at scaled steps (e.g., every other time for
  upsampling by a factor of two) follwed by interpolation. 
  Typically we use bilinear interpolation, but the theory says that
  this makes a stionary signal cyclostationary since the bilinear
  filter is not ``band limited''. The sinc filter has this
  property. Bilinear interpolation is approximately like a sinc
  interpoltation and bicubic interpolation is even closer.}
}
\eat{
 (See the Supplementary Materials for a discussion.)
\sm{I think I got the formula wrong for the downsampling. The covariance in the downsampling case should be tau times the offset.}
}

%\mypar{Generalization to 2D convolutions.} The analysis for 1D case can be easily generalized to 2D. In this case the input would be drawn for a 2D stationary distribution with $K_x(t_1, t_2)$ where $t_1$ and $t_2$ indicate positions in a 2D. 
%The form of the recursion can also be easily derived from Theorem~\ref{th:gp} for example since the recursion only depends on the covaraince function at the same offset. 
%\ds{I put the 2D generalization in the previous section and deleted this one. OK?}

\mypar{Skip connections.} Modern convolutional networks have skip connections where outputs from two layers are added $Z = X + Y$ or concatenated $Z = [X; Y]$. In both cases if $X$ and $Y$ are stationary GPs so is $Z$. See~\cite{garriga2018deep} for a discussion.

\section{Bayesian Inference for Deep Image Prior}
\begin{figure}
    \centering
       \includegraphics[width=1.0\linewidth]{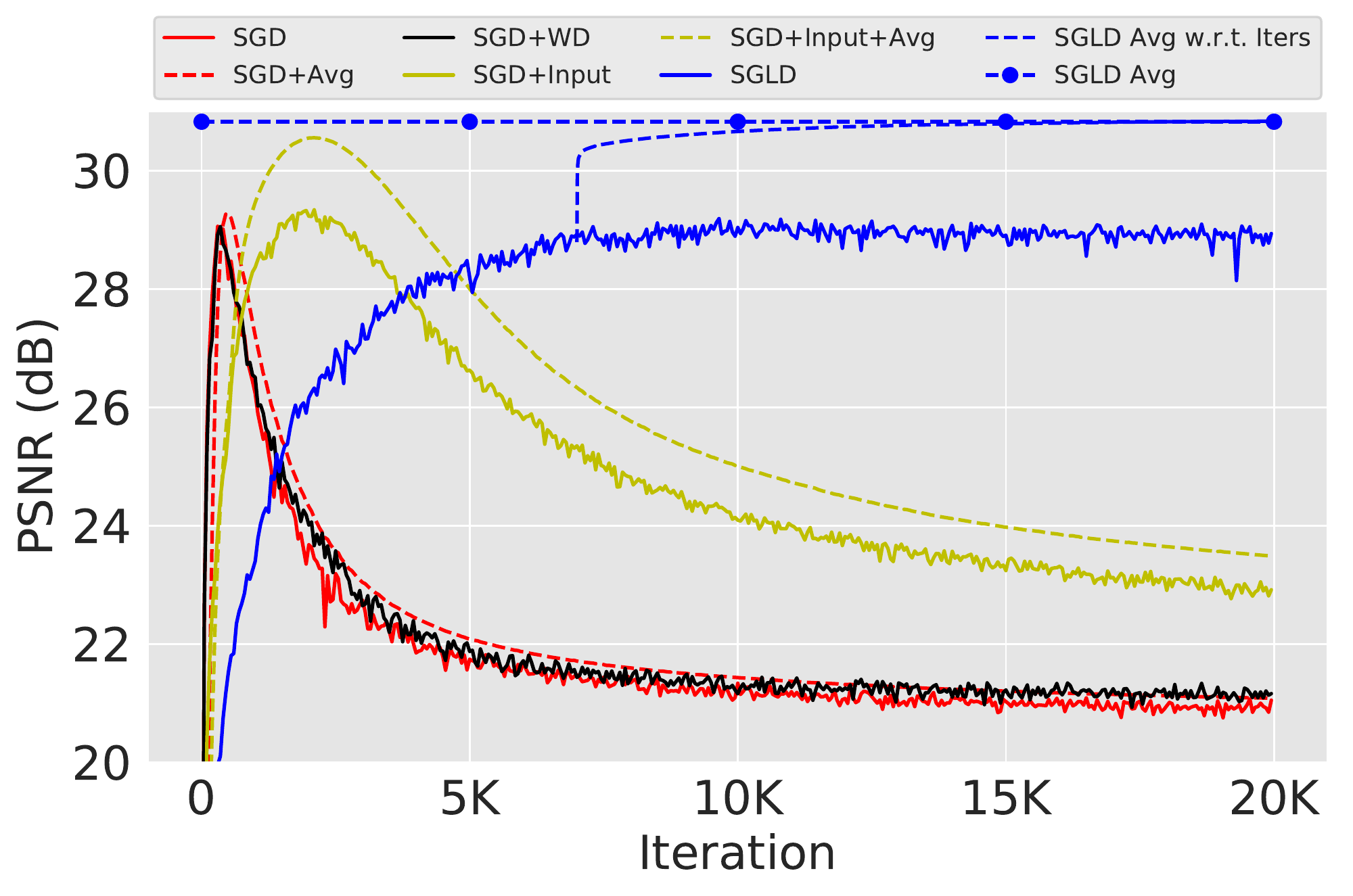}
    \caption{The PSNR curve for different learning methods on the ``peppers" image of Figure~\ref{fig:splash-figure}. The SGD and its variants use early stopping to avoid overfitting. MAP inference by adding a prior term (WD: weight decay) shown as the black curve doesn't avoid overfitting. 
    Moving averages (dashed lines) and adding noise to the input improves performance. By contrast, samples from SGLD after \emph{``burn-in"} remains stable and the posterior mean improves over the highest PSNR of the other approaches.}
    \label{fig:PSNR curve}
    \vspace{-0.2in} 
\end{figure}

Let's revisit the deep image prior for a denoising task. 
Given an noisy image $\hat{y}$ the deep image prior solves 
$$
    \min_{\theta, x} ||\hat{y} - f(x, \theta)||_2^2, 
$$
where $x$ is the input and $\theta$ are the parameters of an appropriately chosen convolutional network. Both $x$ and $\theta$ are initialized randomly from a prior distribution. 
Optimization is performed using stochastic gradient descent (SGD) over $x$ and $\theta$ (optionally $x$ is kept fixed) and relying on early stopping to avoid overfitting (see Figures~\ref{fig:splash-figure} and~\ref{fig:PSNR curve}). The denoised image is obtained as $y^* = f(x^*, \theta^*)$.

The inference procedure can be interpreted as a maximum likelihood estimate (MLE) under a Gaussian noise model: $\hat{y} = y + \epsilon$, where $\epsilon = N(0, \sigma_n^2 \mathbb{I})$.
Bayesian inference suggests we add a suitable prior $p(x, \theta)$ over the parameters and reconstruct the image by \emph{integrating} the posterior, to get
$
    y^* = \int p(x, \theta \mid \hat{y}) f(x, \theta) dx d\theta,
$
The obvious computational challenge is computing this posterior average. An intermediate option is maximum \emph{a posteriori} (MAP) inference where the {\rm argmax} of the posterior is used.  However both MLE and MAP do not capture parameter uncertainty and can overfit to the data.  
\eat{
Bayesian inference suggests we add suitable priors over the parameters $w = \{x, \theta\} $ and integrate the posterior: 
$
    y^* = \int P(w; \hat{y}) y(w) dw,
$
where $P(w; y) \propto p(y; w) p(w)$ and $y(w) = f(x, \theta)$. An intermediate option is Maximum A-Posteriori (MAP) inference where the {\rm argmax} of the posterior is use.  However both MLE and MAP do not capture parameter uncertainty and can overfit to the data.
}

In standard MCMC the integral is replaced by a sample average of a Markov chain that converges to the true posterior. However convergence with MCMC techniques is generally slower than backpropagation for deep networks. 
Stochastic gradient Langevin dyanamics (SGLD)~\cite{welling2011bayesian} provides a general framework to derive an MCMC sampler from SGD by injecting Gaussian noise to the gradient updates. Let $w = (x, \theta)$. The SGLD update is:
\begin{equation}
\begin{split}    
\Delta_w &= \frac{\epsilon}{2}\Big( \nabla_w \log p(\hat{y} \mid w) + \nabla_w \log p(w) \Big) + \eta_t \\
\eta_t &\sim N(0, \epsilon).
\end{split}
\end{equation}
where $\epsilon$ is the step size. Under suitable conditions, \eg, $\sum \epsilon_t = \infty$ and $\sum \epsilon_t^2 < \infty$ and others, it can be shown that $w_1, w_2, \ldots$ converges to the posterior distribution. The log-prior term is implemented as weight decay.

Our strategy for posterior inference with the deep image prior thus adds Gaussian noise to the gradients at each step to estimate the posterior sample averages after a \emph{``burn in"} phase.
As seen in Figure~\ref{fig:splash-figure}(a), due to the Gaussian noise in the gradients, the MSE with respect to the noisy image does not go to zero, and converges to a value that is close to the noise level as seen in Figure~\ref{fig:splash-figure}(b).
It is also important to note that MAP inference alone doesn't avoid overfitting.
Figure~\ref{fig:PSNR curve} shows a version where weight decay is used to regularize parameters, which also overfits to the noise.
Further experiments with inference procedures for denoising are described in Section~\ref{sec:natural}.

%% file: experiments.tex
\section{Experiments}
\def\swidthone{1.0\linewidth}
\def\swidthtwo{0.45\linewidth}
\def\swidththree{0.32\linewidth}
\def\swidthfour{0.22\linewidth}
\def\swidthfive{0.160\linewidth}
\def\swidthsix{0.155\linewidth}
\def\swidthseven{0.150\linewidth}
\vspace{-2pt}

%We organized our experiments as follows: Section \ref{sec:toy_exp} illustrates how the input distribution and network architecture influences the stationary GP kernel and illustrate posterior inference using SGLD on a number of toy 1D examples. Section \ref{sec: denoising} and \ref{sec: inpainting} present the results image denoising and inpainting task where we compare the proposed inference techniques to prior work~\cite{ulyanov18deep}.

\subsection{Toy examples} \label{sec:toy_exp}
We first study the effect of the architecture and input distribution on the covariance function of the stationary GP using 1D convolutional networks.
We consider two architectures: (1) AutoEncoder: where $d$ conv + downsampling blocks are followed by $d$ conv + upsampling blocks, and (2) Conv: where convolutional blocks without any upsampling or downsampling. We use ReLU non-linearity after each conv layer in both cases.
We also vary the input covariance $K_x$. Each channel of $X$ is first sampled iid from a zero-mean Gaussian with a variance $\sigma^2$.
A simple way to obtain inputs with a spatial covariance $K_x$ equal to a Gaussian with standard deviation $\sigma$ is to then spatially filter channels of $X$ with a Gaussian filter with standard deviation $\sqrt{2}\sigma$.

\begin{figure*}[!ht]
    \centering
    \setlength{\tabcolsep}{1pt}
    \begin{tabular}{cccc}
    \includegraphics[width=0.23\linewidth]{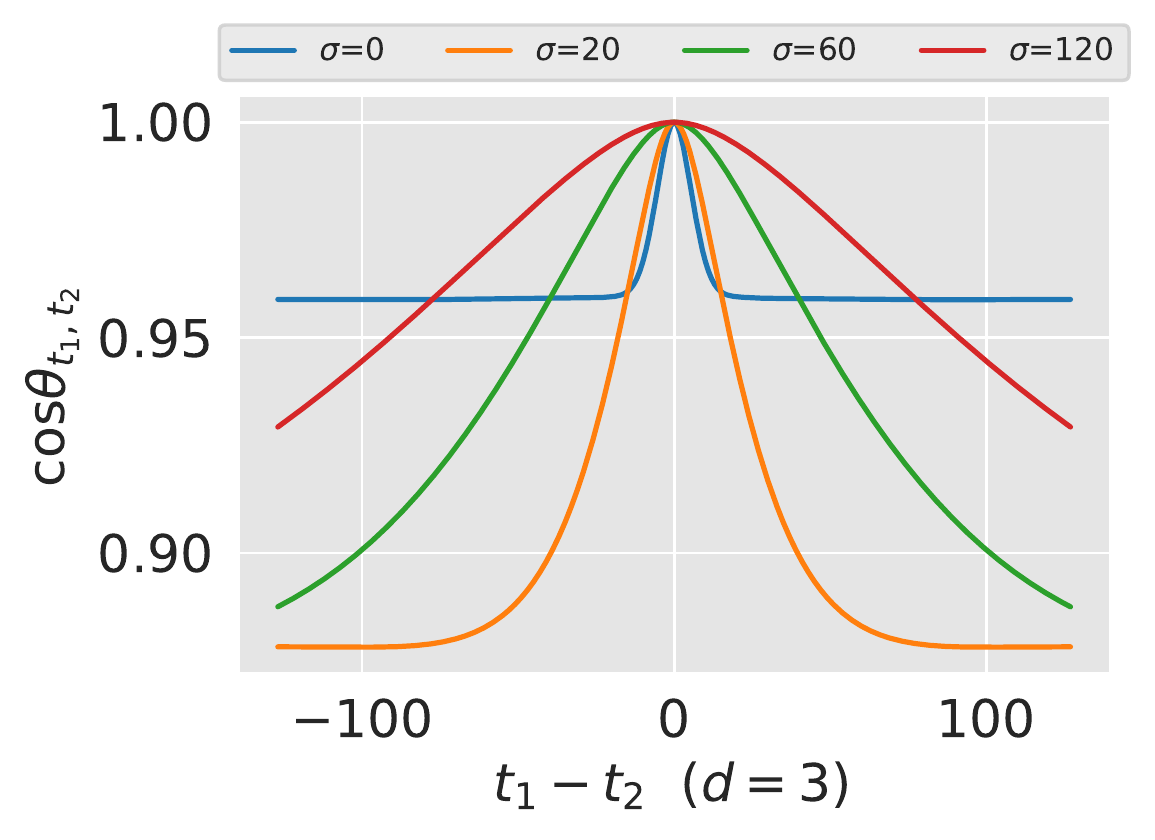} &
    \includegraphics[width=0.235\linewidth]{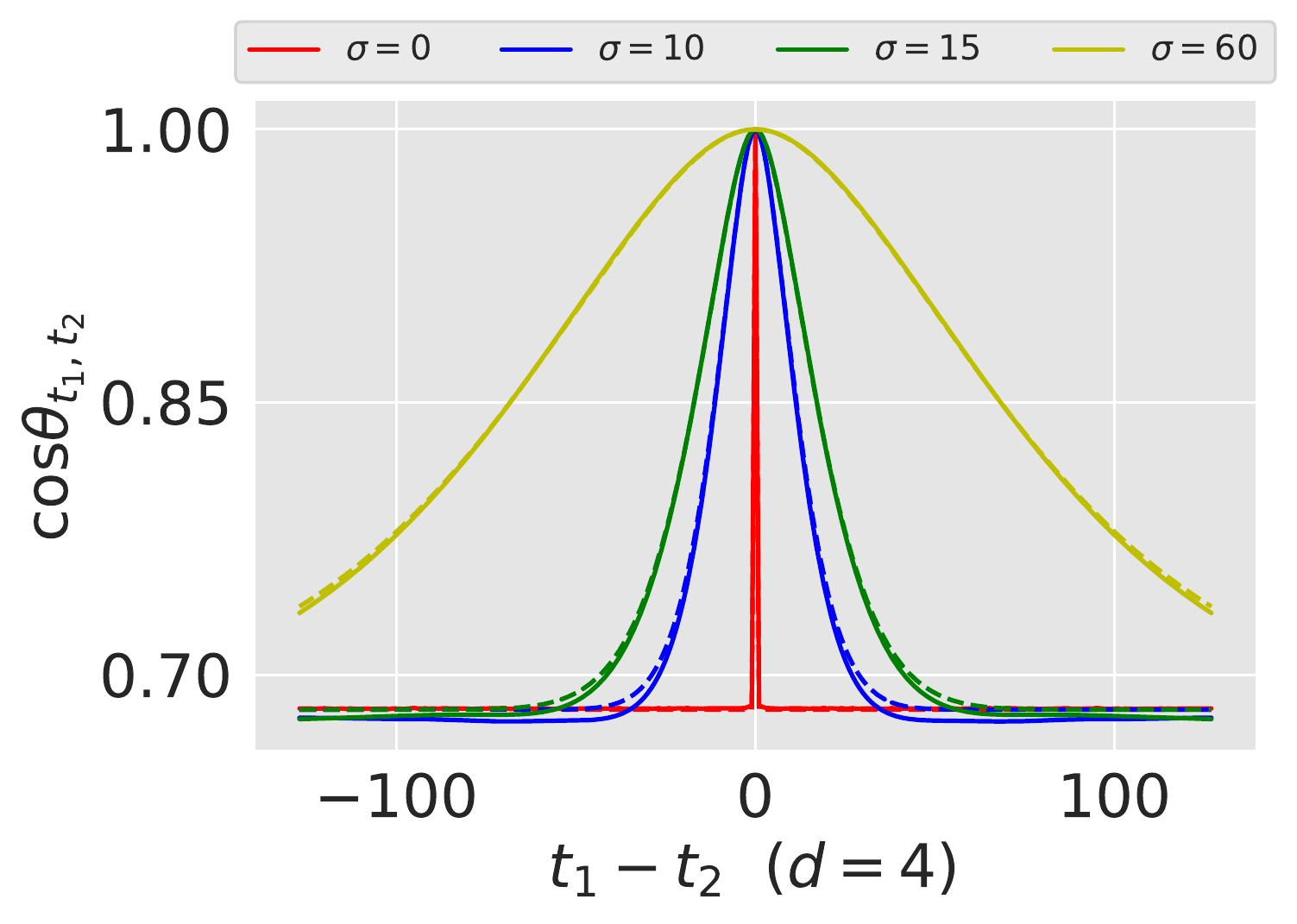} &
    \includegraphics[width=0.22\linewidth]{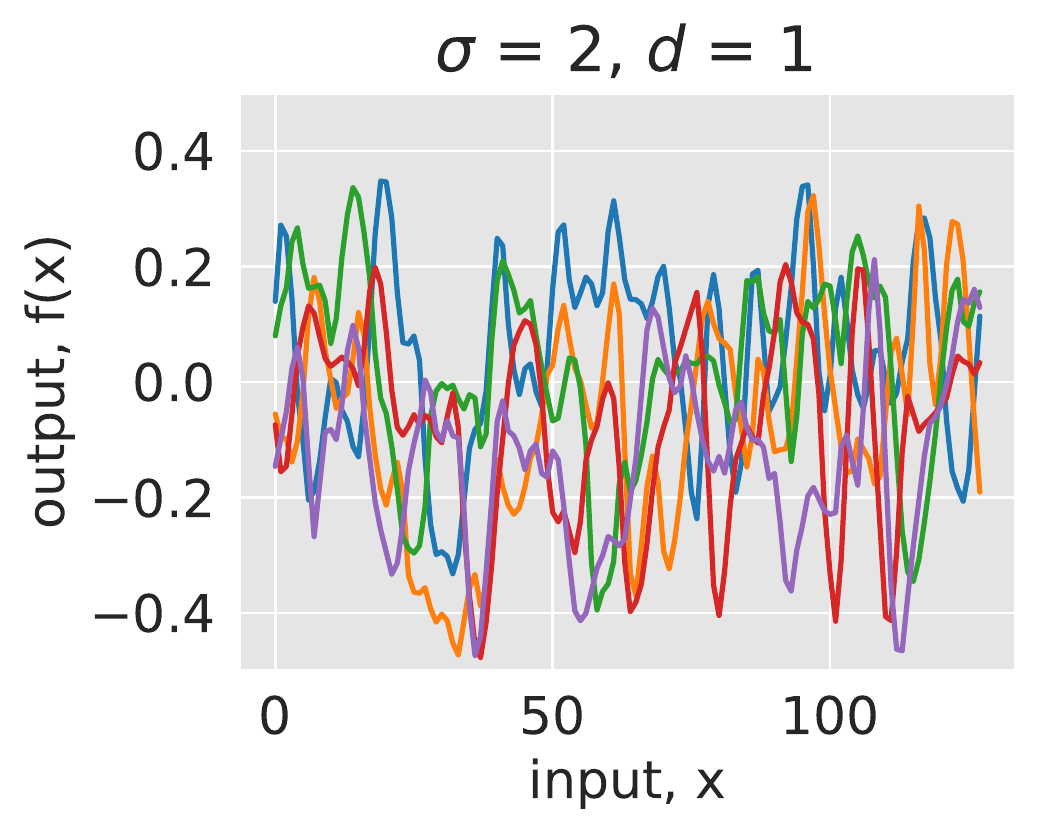} &
    \includegraphics[width=0.22\linewidth]{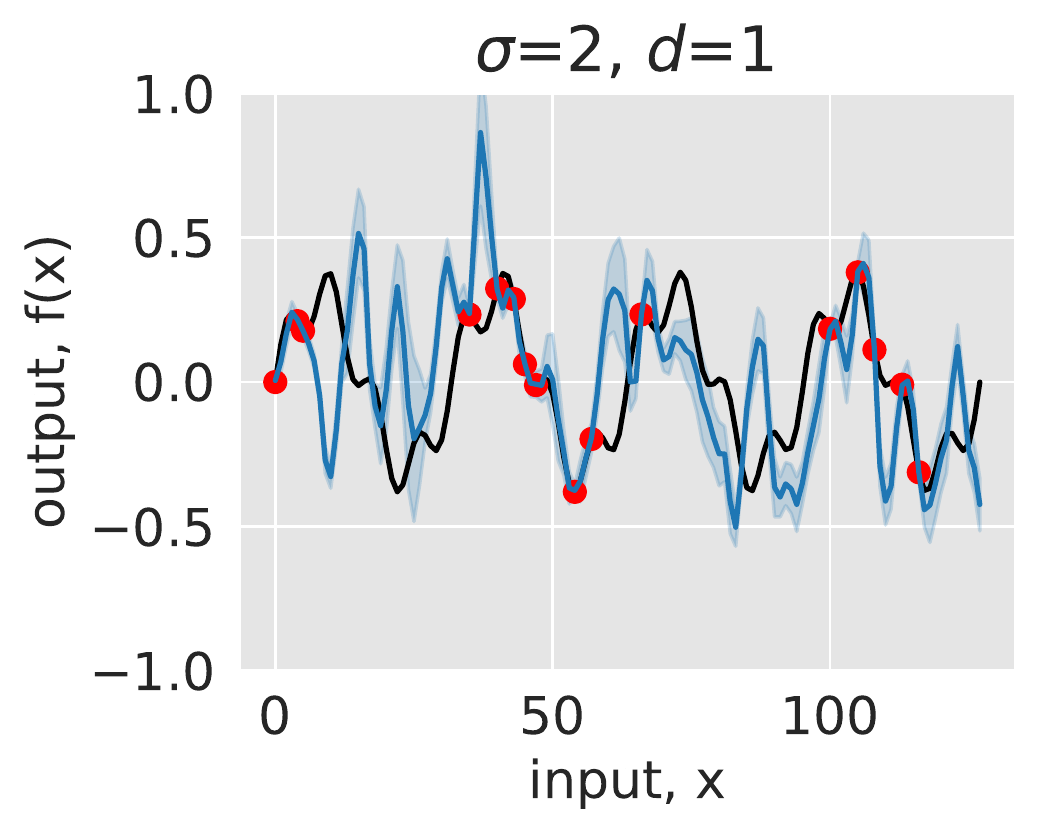} \\
    \includegraphics[width=0.23\linewidth]{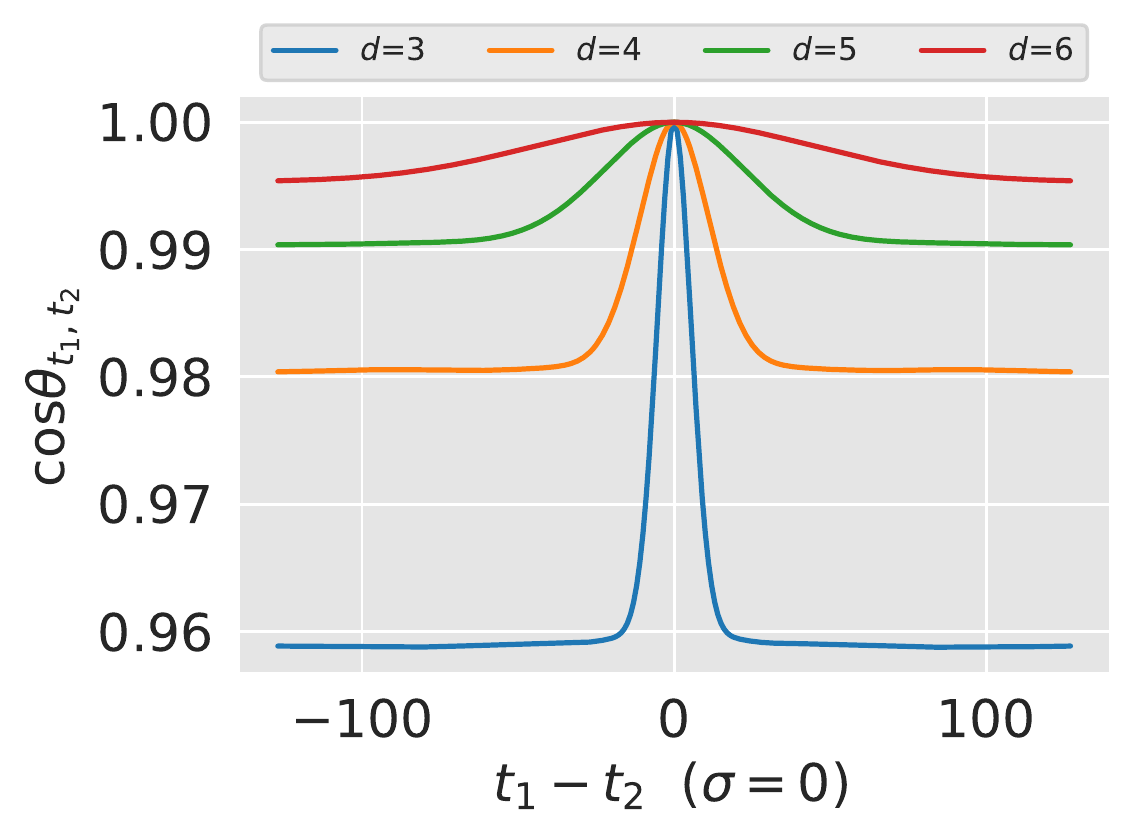} &
    \includegraphics[width=0.23\linewidth]{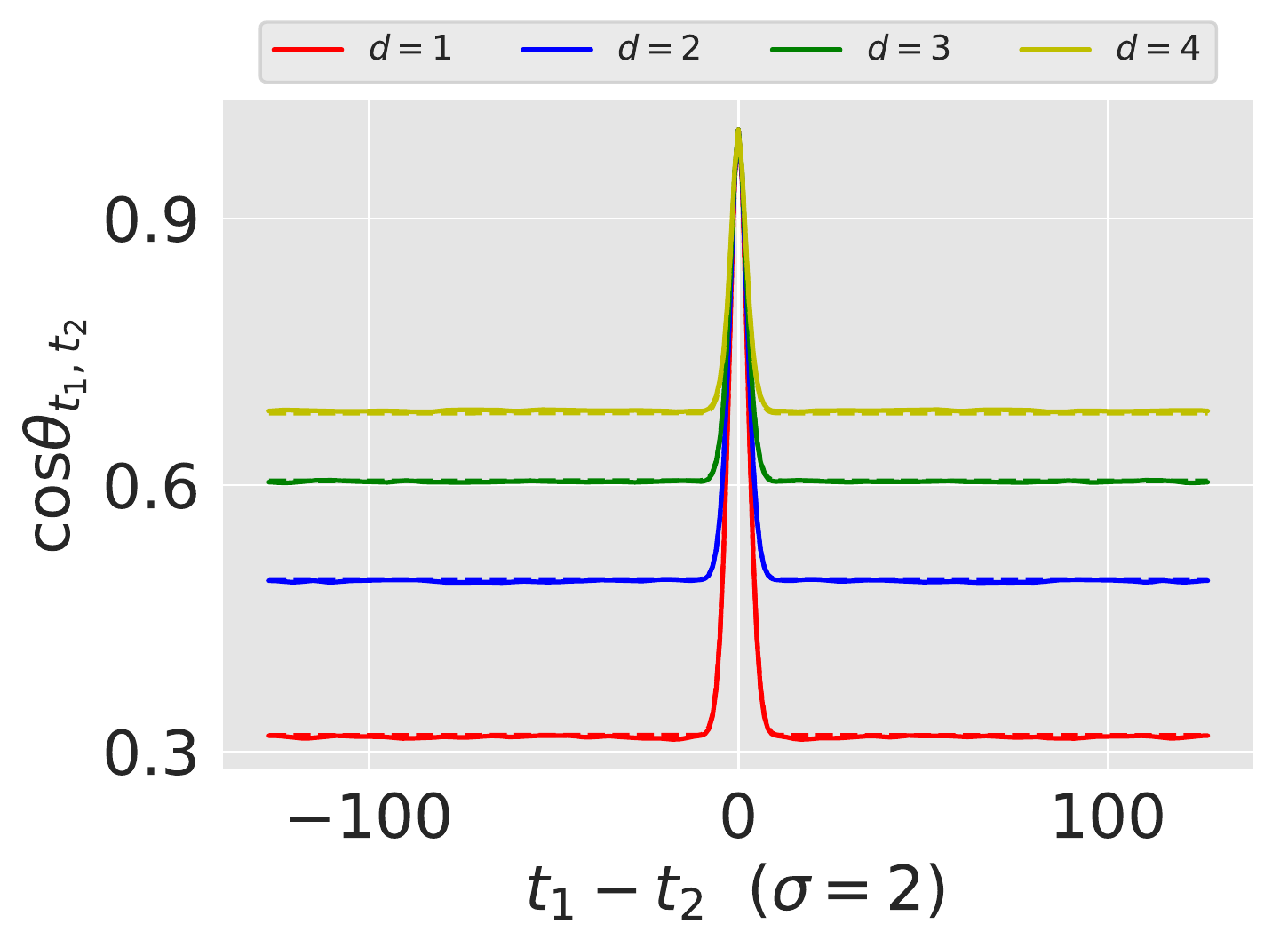} &
    \includegraphics[width=0.22\linewidth]{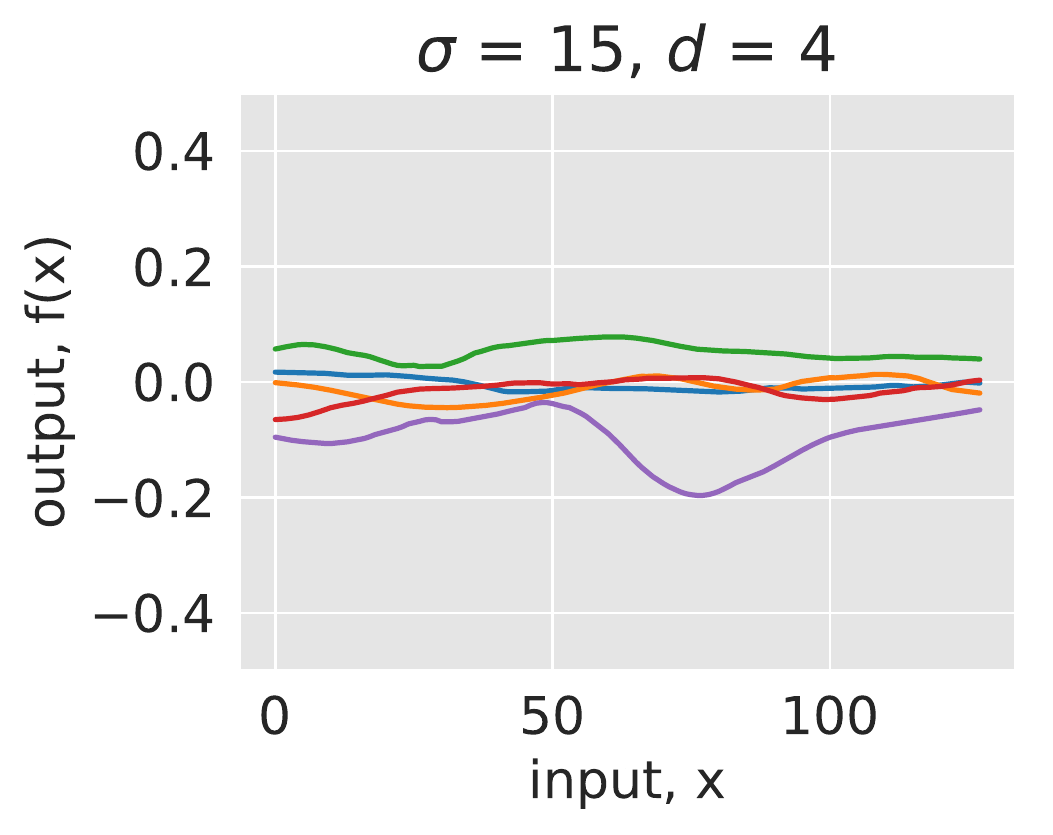} &
     \includegraphics[width=0.22\linewidth]{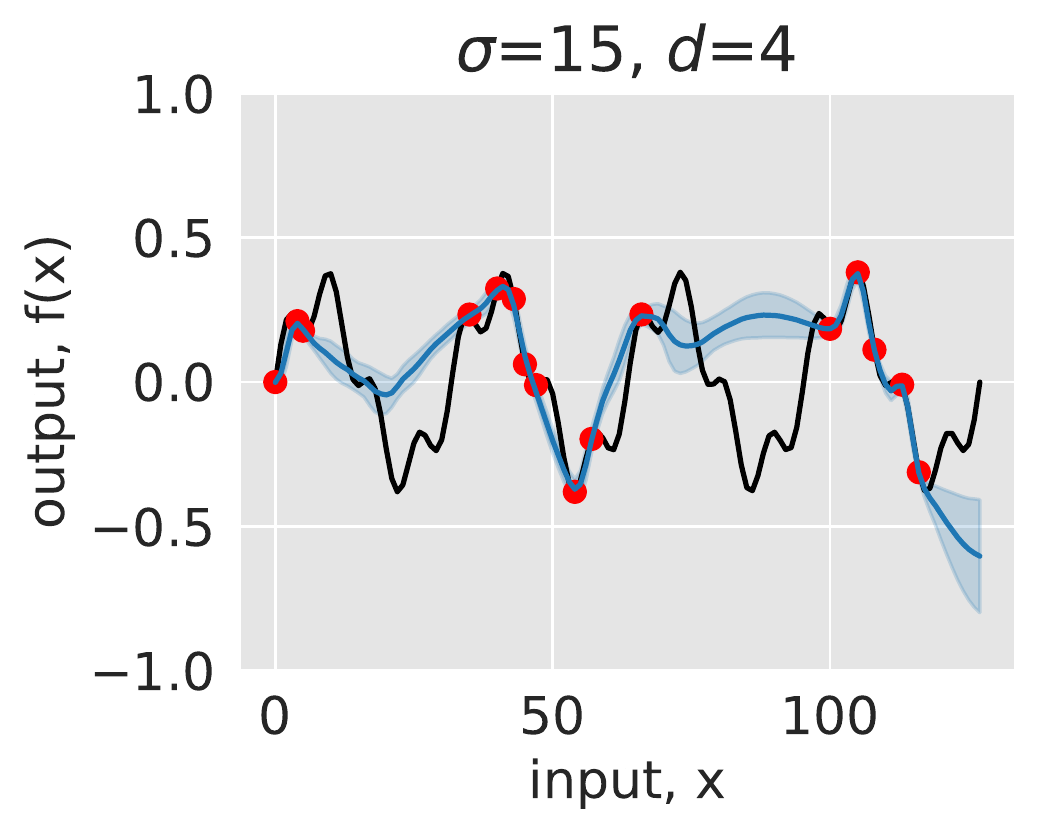} \\
     (a) $\cos \theta_{t_1, t_2}$ (AE) & (b) \new{$\cos \theta_{t_1, t_2}$ (Conv)} & (c) Prior & (d) Posterior \\
     \end{tabular}
    \caption{\small \textbf{Priors and posterior with 1D convolutional networks}. The covariance function $\cos \theta_{t_1, t_2} =  K(t_1-t_2)/K(0)$ for the \textbf{(a)} AutoEncoder and \textbf{(b)} Conv architectures estimated empirically for different values of depth and input covariance. For the Conv architecture we also compute the covariance function analytically using recursion in Equation~\ref{eq:relu-recurrence} shown as \new{dashed lines} in panel \textbf{(b)}. The empirical estimates were obtained with networks with \new{$256$} filters. The agreement is quite good for small values of sigma. For larger offsets the convergence towards a Gaussian is approximate.
    Panel \textbf{(c)} shows samples from the prior of the Conv architecture with two different configurations, and panel \textbf{(d)} shows the posterior means and variances estimated using SGLD.}
    \label{fig:1dtoy}
    \vspace{-0.2in}
\end{figure*}

Figure~\ref{fig:1dtoy} shows the covariance function
$
    \cos \theta_{t_1, t_2} = K_z(t_1-t_2)/K_z(0),
$
induced by varying the $\sigma$ and depth $d$ of the two architectures (Figure~\ref{fig:1dtoy}a-b).
We empirically estimated the covariance function by sampling many networks and inputs from the prior distribution.
The covariance function for the convolutional-only architecture is also calculated using the recursion in Equation~\ref{eq:relu-recurrence}.
For both architectures increasing $\sigma$ and $d$ introduce longer-range spatial covariances.
For the auto-encoder upsampling induces longer-range interactions even when $\sigma$ is zero shedding some light on the role of upsampling in the deep image prior.
Our network architectures have 128 filters, even so, the match between the empirical covariance and the analytic one is quite good as seen in Figure~\ref{fig:1dtoy}(b).

% This was estimated empirically by sampling many networks and inputs from the prior distribution.
% For both architectures increasing $\sigma$ and $d$ introduce longer-range spatial covariances.
% For the auto-encoder upsampling induces longer-range interactions even when $\sigma$ is zero shedding some light on the role of upsampling in the deep image prior.
% The covariance function for the convolutional-only architecture is shown in Figure~\ref{fig:1dtoy}(b) can be calculated using the recursion in Lemma~\ref{lemma:relu} which closely matches the empirical results.
% \todo{show this}.

Figure~\ref{fig:1dtoy}(c) shows samples drawn from the prior of the convolutional-only architecture.
%In this case the $K_z$ can be computed analytically by recursively applying the formula in equation~\ref{eq:relu-recurrence} stating with $K_x$ for the input.
Figure~\ref{fig:1dtoy}(d) shows the posterior mean and variance with SGLD inference where we randomly dropped 90\% of the data from a 1D signal.
Changing the covariance influences the mean and variance which is qualitatively similar to choosing the scale of stationary kernel in the GP: larger scales (bigger input $\sigma$ or depth) lead to smoother interpolations.

\subsection{Natural images}
\label{sec:natural}

Throughout our experiments we adopt the network architecture reported in \cite{ulyanov18deep} for image denoising and inpainting tasks for a direct comparison with their results.
These architectures are 5-layer auto-encoders with skip-connections and each layer contains 128 channels.
We consider images from the standard image reconstruction datasets~\cite{dabov2007image, heide2015fast}.
For inference we use a learning rate of $0.01$ for image denoising and $0.001$ for image inpainting. We compare the following inference schemes:

\begin{enumerate}
\setlength{\itemsep}{0pt}
\item \textbf{SGD+Early:} Vanilla SGD with early stopping.
\item \textbf{SGD+Early+Avg:} Averaging the predictions with exponential sliding window of the vanilla SGD.
\item \textbf{SGD+Input+Early:} Perturbing the input $x$ with an additive Gaussian noise with mean zero and standard deviation $\sigma_p$ at each learning step of SGD.
\item \textbf{SGD+Input+Early+Avg:} Averaging the predictions of the earlier approach with a exponential window.
\item \textbf{SGLD:} Averaging after burn-in iterations of posterior samples with SGLD inference.
\end{enumerate}

We manually set the stopping iteration in the first four schemes to one with essentially the best reconstruction error --- note that this is an oracle scheme and cannot be implemented in real reconstruction settings.
For image denoising task, the stopping iteration is set as 500 for the first two schemes, and 1800 for the third and fourth methods. For image inpainting task, this parameter is set as 5000 and 11000 respectively.

The third and fourth variants were described in the supplementary material of~\cite{ulyanov18deep} and in the released codebase. We found that injecting noise to the input during inference consistently improves results. However, as observed in~\cite{ulyanov18deep}, regardless of the noise variance $\sigma_p$, the network is able to drive the objective to zero, \ie, it overfits to the noise. This is also illustrated in Figure \ref{fig:splash-figure} (a-b).

Since the input $x$ can be considered as part of the parameters, adding noise to the input during inference can be thought of as approximate SGLD.
It is also not beneficial to optimize $x$ in the objective and is kept constant (though adding noise still helps).
SGLD inference includes adding noise to all parameters, $x$ and $\theta$, sampled from a Gaussian distribution with variance scaled as the learning rate $\eta$, as described in Equation~\ref{eq:var_scale}.
We used 7K burn-in iterations and 20K training iterations for image denoising task, 20K and 30K for image inpainting tasks. Running SGLD longer doesn't improve results further.
The weight-decay hyper-parameter for SGLD is set inversely proportional to the number of pixels in the image and equal to 5e-8 for a 1024$\times$1024 image.
For the baseline methods, we did not use weight decay, which, as seen in Figure~\ref{fig:PSNR curve}, doesn't influence results for SGD.

\subsubsection{Image denoising} \label{sec: denoising}

We first consider the image denoising task using various inference schemes. Each method is evaluated on a standard dataset for image denoising~\cite{dabov2007image}, which consists of 9 colored images corrupted with noise of $\sigma = 25$ .

Figure \ref{fig:PSNR curve} presents the peak signal-to-noise ratio (PSNR) values with respect to the clean image over the optimization iterations.
This experiment is on the ``peppers" image from the dataset as seen in Figure~\ref{fig:splash-figure}.
The performance of SGD variants (red, black and yellow curves) reaches a peak but gradually degrades.
By contrast, samples using SGLD (blue curves) are stable with respect to PSNR, alleviating the need for early stopping.
SGD variants benefit from exponential window averaging (dashed red and yellow lines), which also eventually overfits.
Taking the posterior mean after burn in with SGLD (dashed blue line) consistently achieves better performance. The posterior mean at 20K iteration (dashed blue line with markers) achieves the best performance among the various inference methods.

Figure \ref{fig:denoising} shows a qualitative comparison of SGD with early stopping to the posterior mean of SGLD, which contains fewer artifacts. \new{More examples are available in the \newaxv{Appendix}.} Table \ref{tab:denoising} shows the quantitative comparisons between the SGLD and the baselines. We run each method 10 times and report the mean and standard deviations. SGD consistently benefits from perturbing the input signal with noise-based regularization,
%(\ie, SGD with and without input noise)
and from moving averaging.
%(\ie, SGD with/without Avg).
However, as noted, these methods still have to rely on early stopping, which is hard to set in practice.
By contrast, SGLD  outperforms the baseline methods across all images.
Our reported numbers (SGD + Input + Early + Avg) are similar to the single-run results reported in prior work (30.44 PSNR compared to ours of 30.33$\pm$0.03 PSNR.)
SGLD improves the average PNSR to 30.81.
As a reference, BM3D~\cite{dabov2007image} obtains an average PSNR of 31.68.

\begin{figure}
    \centering
    \setlength{\tabcolsep}{1pt}
    \begin{tabular}{ccc}
    \includegraphics[width=\swidththree]{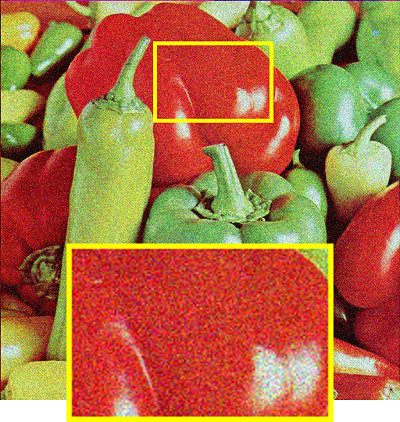} &
    \includegraphics[width=\swidththree]{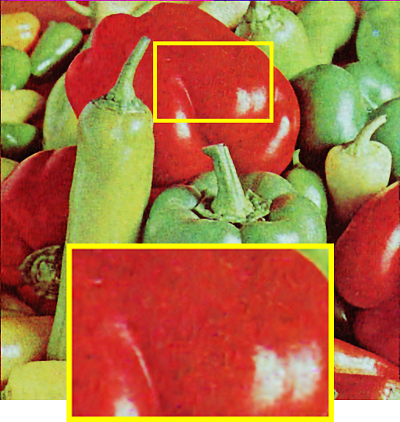} &
    \includegraphics[width=\swidththree]{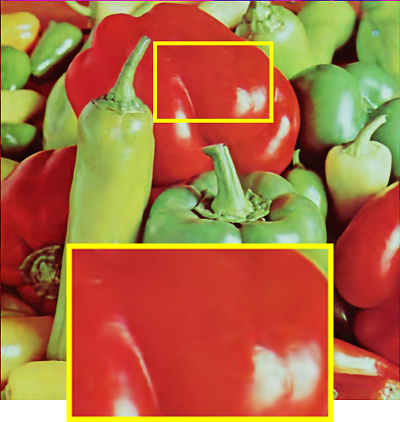} \\
    Input & SGD (28.38) & SGLD (30.82) \\
    \end{tabular}
    \caption{\textbf{\new{Image denoising results.}} Denoising the
      input noisy image with SGD and SGLD inference.}
    \label{fig:denoising}
    \vspace{-15pt}
\end{figure}

\eat{
\begin{figure}[!h]
    \centering
    \setlength{\tabcolsep}{1pt}
    \begin{tabular}{ccc}
    \includegraphics[width=\swidththree]{figs/Denoising/2-1.png}&
    \includegraphics[width=\swidththree]{figs/Denoising/4-1.png}&
    \includegraphics[width=\swidththree]{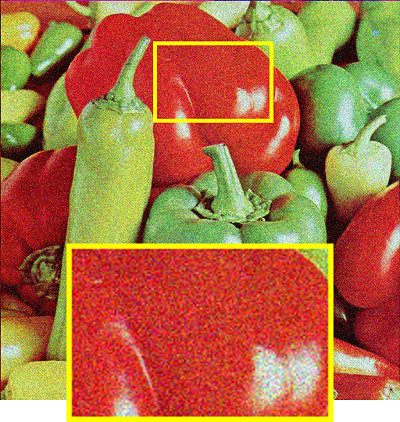} \\
    \includegraphics[width=\swidththree]{figs/Denoising/2-2.png} &
    \includegraphics[width=\swidththree]{figs/Denoising/4-2.png}&
    \includegraphics[width=\swidththree]{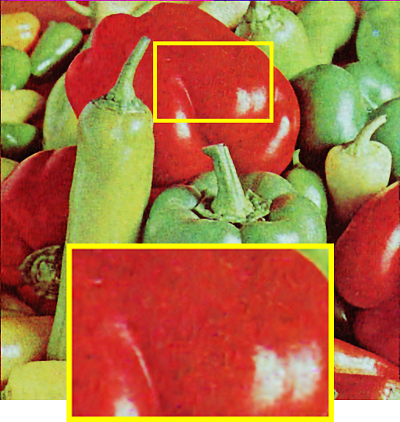} \\
    29.81  & 29.07 & 28.38 \\
    \includegraphics[width=\swidththree]{figs/Denoising/2-3.png} &
    \includegraphics[width=\swidththree]{figs/Denoising/4-3.png}&
    \includegraphics[width=\swidththree]{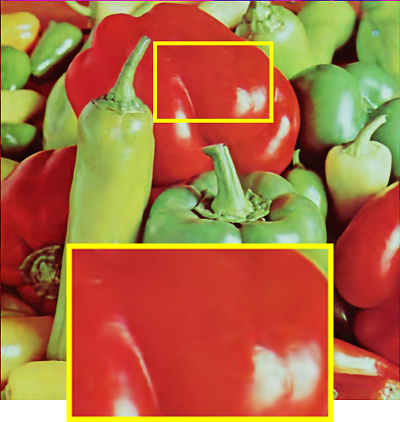} \\
    32.95 & 32.08 & 30.82 \\
    \end{tabular}
    % \caption{}
    \caption{\textbf{Image denoising results.} Denoising the input noisy image (top row) with SGD (middle row) and SGLD (bottom row) inference. The PSNR is indicated below each image.}
    \label{fig:denoising}
\end{figure}
}

\vspace{-10pt}
\subsubsection{Image inpainting} \label{sec: inpainting}
\vspace{-3pt}

For image inpainting we experiment on the same task as~\cite{ulyanov18deep}
%the original paper
where 50\% of the pixels are randomly dropped.
We evaluate various inference schemes on the standard image inpainting dataset~\cite{heide2015fast} consisting of 11 grayscale images.

Table \ref{Table:inpainting} presents a comparison between SGLD and the baseline methods.
Similar to the image denoising task, the performance of SGD is improved by perturbing the input signal and additionally by averaging the intermediate samples during optimization.
SGLD inference provides additional improvements; it outperforms the baselines and improves over the results reported in \cite{ulyanov18deep} from 33.48 to 34.51 PSNR.
Figure \ref{fig:inpainting} shows qualitative comparisons between SGLD and SGD. The posterior mean of SGLD has fewer artifacts than the best result generated by SGD variants.

Besides gains in performance, SGLD provides estimates of uncertainty. This is visualized in Figure \ref{fig:splash-figure}(d). Observe that uncertainty is low in missing regions that are surrounded by areas of relatively uniform appearance such as the window and floor, and higher in non-uniform areas such as those near the boundaries of different object in the image.

\begin{table*}[tb]
\centering
\caption{\textbf{Image denoising task.} Comparison of various inference schemes with the deep image prior for image denoising ($\sigma$=25). Bayesian inference with SGLD avoids the need for early stopping while consistently improves results. Details are described in Section~\ref{sec: denoising}.}
%\vspace{0.1in}
\setlength{\tabcolsep}{4pt}
\begin{tabular}{r|ccccccccc|c}
        \toprule
                     & House & Peppers & Lena & Baboon & F16 & Kodak1 & Kodak2 & Kodak3 & Kodak12 & Average \\
        \midrule
        SGD + Early  & 26.74 & 28.42 & 29.17 & 23.50 & 29.76 & 26.61 & 28.68 & 30.07 & 29.78 & 28.08 \\ [-0.04in]
                     & \scriptsize{$\pm$0.41} & \scriptsize{$\pm$0.22} & \scriptsize{$\pm$0.25} & \scriptsize{$\pm$0.27} & \scriptsize{$\pm$0.49} & \scriptsize{$\pm$0.19} & \scriptsize{$\pm$0.18} & \scriptsize{$\pm$0.33} & \scriptsize{$\pm$0.17} & \scriptsize{$\pm$0.09} \\
        SGD + Early + Avg      & 28.78 & 29.20 & 30.26 & 23.82 & 31.17 & 27.14 & 29.88 & 31.00 & 30.64 & 29.10 \\[-0.04in]
                     & \scriptsize{$\pm$0.35} & \scriptsize{$\pm$0.08} & \scriptsize{$\pm$0.12} & \scriptsize{$\pm$0.11} & \scriptsize{$\pm$0.1} & \scriptsize{$\pm$0.07} & \scriptsize{$\pm$0.12} & \scriptsize{$\pm$0.11} & \scriptsize{$\pm$0.12} & \scriptsize{$\pm$0.05} \\
       SGD + Input  + Early      & 28.18 & 29.21 & 30.17 & 22.65 & 30.57 & 26.22 & 30.29 & 31.31 & 30.66 & 28.81 \\[-0.04in]
                     & \scriptsize{$\pm$0.32} & \scriptsize{$\pm$0.11} & \scriptsize{$\pm$0.07} & \scriptsize{$\pm$0.08} & \scriptsize{$\pm$0.09} & \scriptsize{$\pm$0.14} & \scriptsize{$\pm$0.13} & \scriptsize{$\pm$0.08} & \scriptsize{$\pm$0.12} & \scriptsize{$\pm$0.04} \\
       SGD + Input + Early + Avg  & 30.61 & 30.46 & 31.81 & 23.69 & 32.66 & 27.32 & 31.70 & 32.86 & 31.87 & 30.33 \\[-0.04in]
                     & \scriptsize{$\pm$0.3} & \scriptsize{$\pm$0.03} & \scriptsize{$\pm$0.03} & \scriptsize{$\pm$0.09} & \scriptsize{$\pm$0.06} & \scriptsize{$\pm$0.06} & \scriptsize{$\pm$0.03} & \scriptsize{$\pm$0.08} & \scriptsize{$\pm$0.1} & \scriptsize{$\pm$0.03} \\
        SGLD         & \textbf{30.86} & \textbf{30.82} & \textbf{32.05} & \textbf{24.54} & \textbf{32.90} & \textbf{27.96} & \textbf{32.05} & \textbf{33.29} & \textbf{32.79} & \textbf{30.81 }\\ [-0.04in]
                     & \scriptsize{$\pm$0.61} & \scriptsize{$\pm$0.01} & \scriptsize{$\pm$0.03} & \scriptsize{$\pm$0.04} & \scriptsize{$\pm$0.08} & \scriptsize{$\pm$0.06} & \scriptsize{$\pm$0.05} & \scriptsize{$\pm$0.17} & \scriptsize{$\pm$0.06} & \scriptsize{$\pm$0.08} \\
        \hline
        CMB3D \cite{dabov2007image}   & 33.03 & 31.20	& 32.27	& 25.95	 & 32.78 & 29.13 & 32.44 & 34.54	& 33.76	& 31.68 \\
        \bottomrule
\end{tabular}
\vspace{-0.01in}
\label{tab:denoising}
\end{table*}

\begin{table*}
\centering
\caption{\textbf{Image inpainting task.} Comparison of various inference schemes with the deep image prior for  image inpainting. SGLD estimates are more accurate while also providing a sensible estimate of the variance. Details are described in Section~\ref{sec: inpainting}.}
% \vspace{0.1in}
\setlength{\tabcolsep}{2pt}
\begin{tabular}{r|ccccccccccc|c}
\toprule
Method & Barbara & Boat & House & Lena & Peppers & C.man & Couple & Finger & Hill & Man & Montage & Average \\
\midrule
SGD + Early & 28.48 & 31.54 & 35.34 & 35.00 & 30.40 & 27.05 & 30.55 & 32.24 & 31.37 & 31.32 & 30.21 & 31.23\scriptsize\\ [-0.04in]
            & \scriptsize{$\pm$0.99} & \scriptsize{$\pm$0.23} & \scriptsize{$\pm$0.45} & \scriptsize{$\pm$0.25} & \scriptsize{$\pm$0.59} & \scriptsize{$\pm$0.35} & \scriptsize{$\pm$0.19} & \scriptsize{$\pm$0.16} & \scriptsize{$\pm$0.35} & \scriptsize{$\pm$0.29} & \scriptsize{$\pm$0.82} & \scriptsize{$\pm$0.11} \\
SGD + Early + Avg       & 28.71 & 31.64 & 35.45 & 35.15 & 30.48 & 27.12 & 30.63 & 32.39 & 31.44 & 31.50 & 30.25 & 31.34\\[-0.04in]
              &  \scriptsize{$\pm$0.7} &  \scriptsize{$\pm$0.28} &  \scriptsize{$\pm$0.46} &  \scriptsize{$\pm$0.18} &  \scriptsize{$\pm$0.6} &  \scriptsize{$\pm$0.39} &  \scriptsize{$\pm$0.18} &  \scriptsize{$\pm$0.12} &  \scriptsize{$\pm$0.31} &  \scriptsize{$\pm$0.39} &  \scriptsize{$\pm$0.82} &  \scriptsize{$\pm$0.08} \\
SGD + Input  + Early     & 32.48 & 32.71 & 36.16 & 36.91 & 33.22 & 29.66 & 32.40 & 32.79 & 33.27 & 32.59 & 33.15 & 33.21\\[-0.04in]
            & \scriptsize{$\pm$0.48} & \scriptsize{$\pm$1.12} & \scriptsize{$\pm$2.14} & \scriptsize{$\pm$0.19} & \scriptsize{$\pm$0.24} & \scriptsize{$\pm$0.25} & \scriptsize{$\pm$2.07} & \scriptsize{$\pm$0.94} & \scriptsize{$\pm$0.07} & \scriptsize{$\pm$0.14} & \scriptsize{$\pm$0.46} & \scriptsize{$\pm$0.36} \\
SGD + Input + Early + Avg & 33.18 & 33.61 & 37.00 & 37.39 & 33.53 & 29.96 & 33.30 & 33.17 & 33.58 & 32.95 & 33.80 & 33.77\\[-0.04in]
             & \scriptsize{$\pm$0.45} & \scriptsize{$\pm$0.3} & \scriptsize{$\pm$2.01} & \scriptsize{$\pm$0.14} & \scriptsize{$\pm$0.31} & \scriptsize{$\pm$0.3} & \scriptsize{$\pm$0.15} & \scriptsize{$\pm$0.77} & \scriptsize{$\pm$0.19} & \scriptsize{$\pm$0.16} & \scriptsize{$\pm$0.6} & \scriptsize{$\pm$0.23} \\
SGLD         & \textbf{33.82} & \textbf{34.26} & \textbf{40.13} &\textbf{ 37.73} &\textbf{ 33.97} & \textbf{30.33} & \textbf{33.72} & \textbf{33.41} & \textbf{34.03} & \textbf{33.54} & \textbf{34.65} & \textbf{34.51} \\[-0.04in]
            & \scriptsize{$\pm$0.19} & \scriptsize{$\pm$0.12} & \scriptsize{$\pm$0.16} & \scriptsize{$\pm$0.05} & \scriptsize{$\pm$0.15} & \scriptsize{$\pm$0.15} & \scriptsize{$\pm$0.1} & \scriptsize{$\pm$0.04} & \scriptsize{$\pm$0.03} & \scriptsize{$\pm$0.06} & \scriptsize{$\pm$0.72} & \scriptsize{$\pm$0.08} \\
\hline
Ulyanov \etal \cite{ulyanov18deep} & 32.22 & 33.06 & 39.16 & 36.16 & 33.05 & 29.80 & 32.52 & 32.84 & 32.77 & 32.2 & 34.54 & 33.48 \\
Papyan \etal \cite{papyan2017convolutional} & 28.44 & 31.44 & 34.58 & 35.04 & 31.11 & 27.90 & 31.18 & 31.34 & 32.35 & 31.92 & 28.05 & 31.19\\
\bottomrule
\end{tabular}
\vspace{-0.01in}
\label{Table:inpainting}
\end{table*}

\subsection{\new{Equivalence between GP and DIP}} \label{sec:GP-DIP}

We compare the deep image prior (DIP) and its Gaussian process
(GP) counterpart, both as prior and for posterior inference, and as a function of the
number of filters in the network.
For efficiency we used a U-Net architecture with two downsampling
and upsampling layers for the DIP.

%Figure~\ref{fig:1}(b) for image inpainting. The network architecture and
%image size were chosen for efficiency.

\vspace{-10pt}
\begin{figure}[th]
    \setlength{\tabcolsep}{1pt}
    \centering
    \begin{tabular}{cc}
        \includegraphics[height=0.23\linewidth]{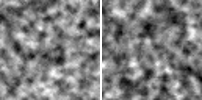} &
        \includegraphics[height=0.23\linewidth]{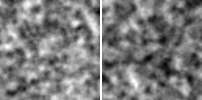}\\
        \small{(a) DIP prior samples} & \small{(b) GP prior samples}\\
       \end{tabular}
       % \caption{\new{Prior samples from DIP and its GP counterpart.}}
       \vspace{-13pt}
    \label{fig:2}
\end{figure}

\begin{figure*} [ht]
    \setlength{\tabcolsep}{1pt}
    \centering
    \begin{tabular}{cccc}
    \includegraphics[width=0.25\linewidth]{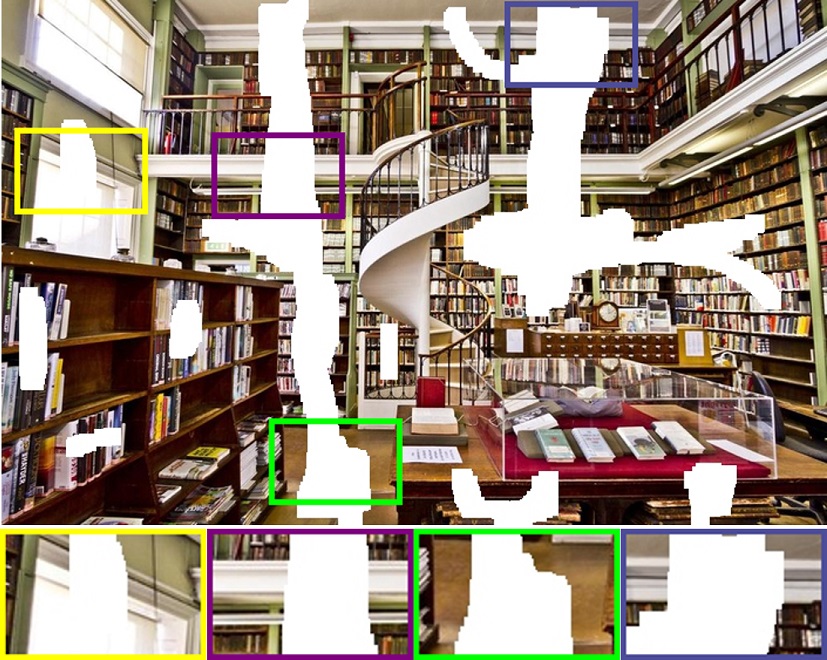} &
    \includegraphics[width=0.25\linewidth]{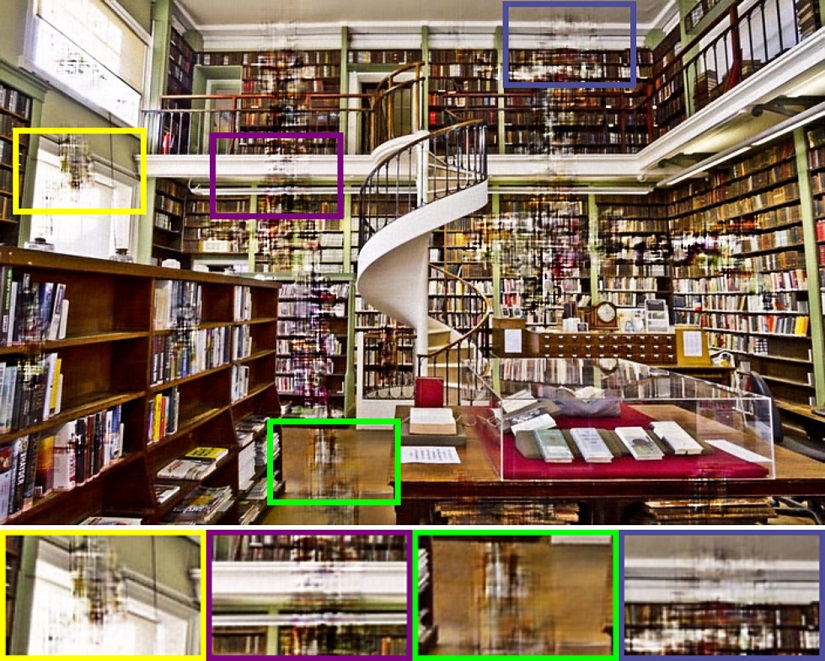} &
    \includegraphics[width=0.25\linewidth]{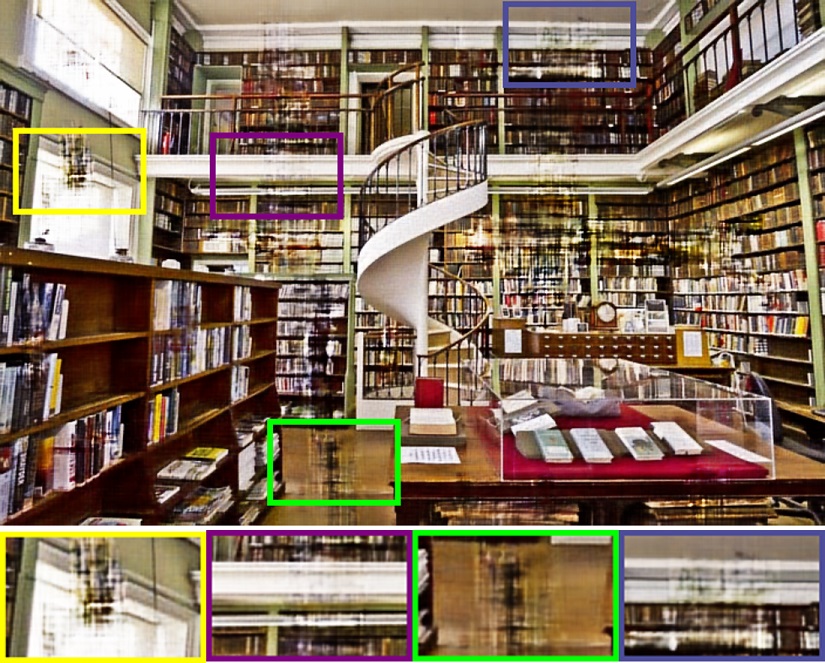} &
        \includegraphics[width=0.25\linewidth]{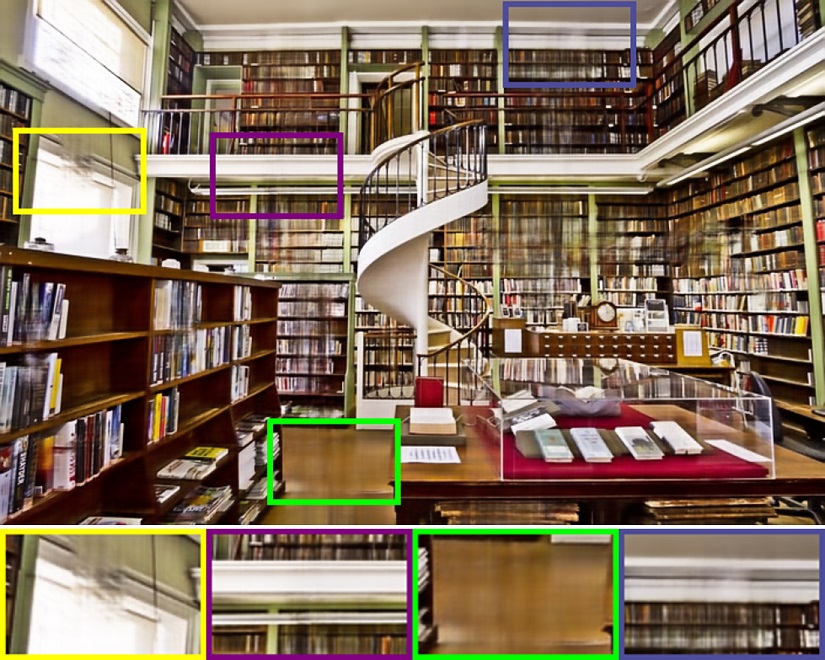} \\
        (a) Input & (b) SGD (19.23 dB)  & (c) SGD + Input (19.59 dB) & (d) SGLD mean (21.86 dB) \\

    \end{tabular}
    \caption{\emph{(Best viewed magnified.)} \textbf{Image inpainting using the deep image prior.} The posterior mean using SGLD (Panel \textbf{(d)}) achieves higher PSNR values and has fewer artifacts than SGD variants. \new{See the \newaxv{Appendix} for more comprisons.}}
    \label{fig:inpainting}
    \vspace{-0.02in}
\end{figure*}

\begin{figure*} [ht]
  \setlength{\tabcolsep}{0.3pt}
  \centering
        \begin{tabular}{cccccc}
        \includegraphics[height=0.160\linewidth]{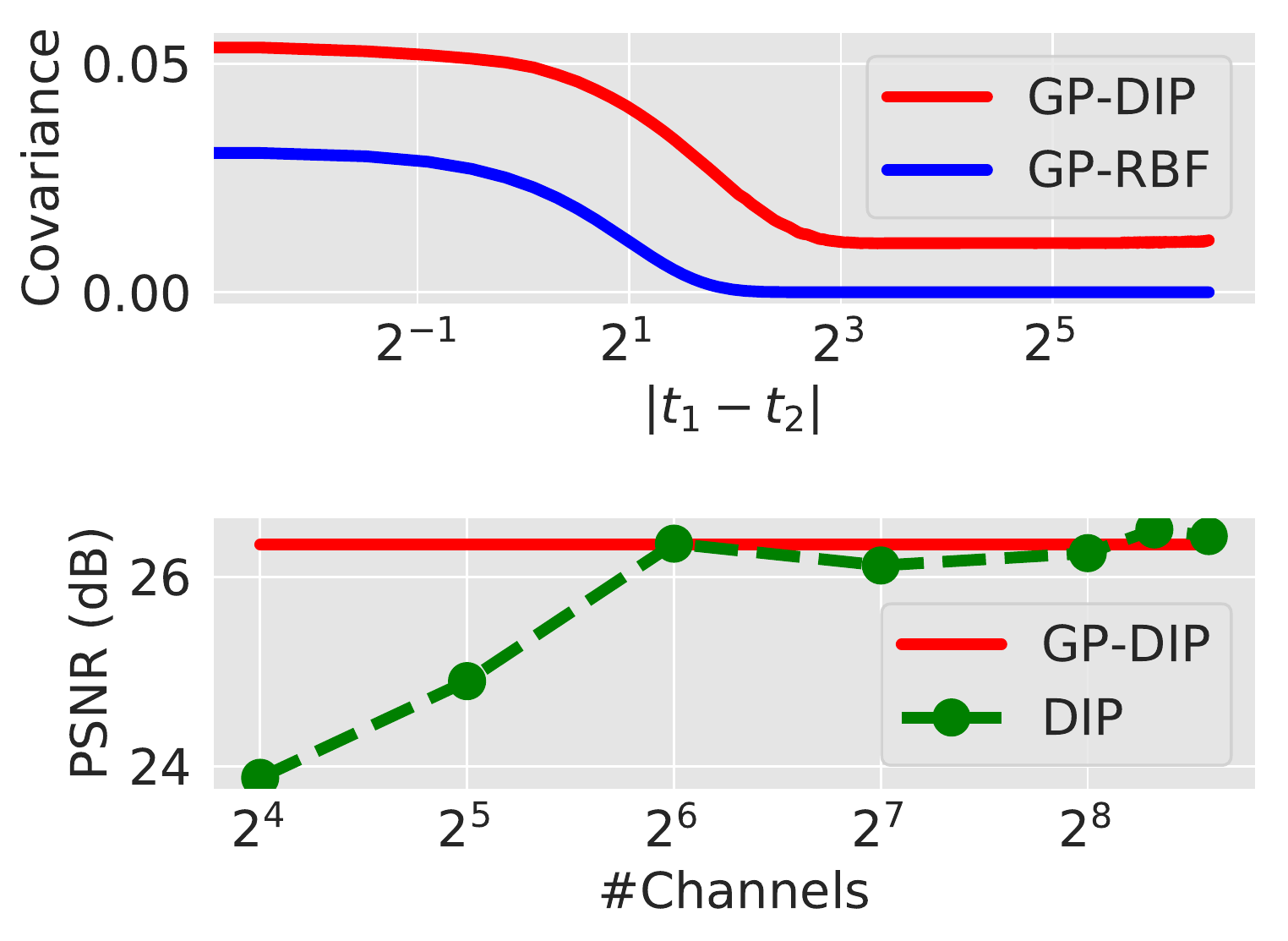} &
        \includegraphics[width=\swidthsix]{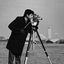} &
        \includegraphics[width=\swidthsix]{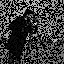} &
        \includegraphics[width=\swidthsix]{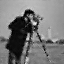} &
        \includegraphics[width=\swidthsix]{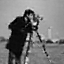} &
        \includegraphics[width=\swidthsix]{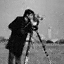} \\
        (a)  & \small{(b) GT}  & \small{(c) Corrupted} &  \small{(d) GP RBF (25.78)}  &  \small{(e) GP DIP (26.34)} &  \small{(f) DIP (26.43)}\\
       \end{tabular}
      % \vspace{-10pt}
    \caption{
    %\new{Inpainting with a Gaussian process (GP) and deep image prior (DIP). \textbf{Top (a)} Comparison of theRBF kernel (learned on (c)) and the stationary DIP kernel. \textbf{Bottom (a)} PSNR of the GP posterior with the DIP kernel and DIP as a function of the number of channels. \textbf{(d-f)} Inpainting results (with PSNR) from GP with the RBF (GP RBF) and DIP (GP DIP) kernel, as well as the deep image prior.}}
     \new{\textbf{Inpainting with a Gaussian process (GP) and deep image prior (DIP).} \textbf{Top (a)} Comparison of the Radial basis function (RBF) kernel with the length scale learned on observed pixels in \textbf{(c)} and the stationary DIP kernel. \textbf{Bottom (a)} PSNR of the GP posterior with the DIP kernel and DIP as a function of the number of channels. DIP approaches the GP performance as the number of channels increases from 16 to 512. \textbf{(d - f)} Inpainting results (with the PSNR values) from GP with the RBF (GP RBF) and DIP (GP DIP) kernel, as well as the deep image prior. The DIP kernel is more effective than the RBF.}}
    \label{fig:1}
\end{figure*}

% Figure~\ref{fig:2}
The above figure shows two samples each drawn from the DIP (\new{with 256 channels per layer}) and GP
with the \emph{equivalent} kernel. The
samples are nearly identical suggesting that the
characterization of the DIP as a stationary GP also holds for 2D
signals.
% \sm{How many channels were used?}
%We found that draws from the DIP were similar to draws from the
%equivalent stationary GP across the range of architectures used in our
%experiments.
% We found that samples drawn the GPs with the DIP kernel computed analytically (or empirically) to be in similar to the samples drawn from the DIP as seen Figure \ref{fig:2}. Note that the samples from the GP and DIP prior were similar across a wide range of architectures.
% For the posterior inference we ran several experiments on image inpainting on a 64$\times$64 image shown in Fig~\ref{fig:1} (b). The size was chosen for efficiency.
% We evaluated the performance of deep image prior (DIP) with a U-Net
% architecture with two downsampling and upsampling layers.
Next, we compare the DIP and GP on an inpainting task shown in
Figure~\ref{fig:1}. The image size here is 64$\times$64.
Figure~\ref{fig:1} top (a) shows the RBF and DIP kernels as a function of the offset.
The DIP kernels are heavy tailed in comparison to Gaussian with
support at larger length scales.
% This also clarifies a issue raised by R3 that the auto-correlation function of natural image are heavy-tailed.
 Figure~\ref{fig:1} bottom (a) shows the performance (PSNR) of the DIP as a
 function of the number of channels from 16 to 512 in each layer of
 the U-Net, \new{as well as of a GP with the limiting DIP
 kernel.}
% \ds{The figure does not appear to have PSNR of GP-RBF.}
% \sm{Not sure who added the blue text, but Figure 6(a) only compares
%  the GP-DIP and DIP. The PNSR with the GP kernel is shown in the
%  other panels.}
 The PSNR of the DIP approaches the GP as the number of
 channels increases suggesting that for networks of this size 256
 filters are enough for the asymptotic GP behavior.
% For deeper networks we were unable to verify at what point does the DIP posterior match GPs because we couldn't reasonably increase the number of channels to a very large number due to GPU memory constraints.
% The estimated images for GP with the RBF and DIP kernels as well as the DIP is shown in Fig \ref{fig:1} (d-f).
 Figure~\ref{fig:1}~(d-e) show that a GP with the DIP kernel is more effective
than one with the RBF kernel, suggesting that the long-tail DIP kernel
is better suited for modeling natural images.

While DIPs are asymptotically GPs, the SGD optimization may be
preferable because GP inference is expensive for high-resolution
images. The memory usage is $O(n^2)$ and running time is $O(n^3)$ for
exact inference where $n$ is the number of pixels (\eg, a
500$\times$500 image requires 233 GB memory). The DIP's memory
footprint, on the other hand, scales linearly with the number of
pixels, and inference with SGD is practical and efficient.
This emphasizes the importance of SGLD, \new{which addresses the drawbacks of
vanilla SGD and makes the DIP more robust and effective.} % \ds{slight rewording}
\new{
Finally, while we showed that the prior distribution induced by the DIP
is asymptotically a GP and the posterior estimated by SGD or SGLD matches the GP posterior for
small networks, it remains an open question if the posterior matches the GP posterior for deeper
networks.
}

%Some prior work~\cite{novak2018bayesian} has noted that the posteriors
%are indeed different and we speculate that this might be due to an
%\emph{implicit bias} induced by the optimization process (\eg,
%~\cite{gunasekar2018characterizing,gunasekar2018implicit}).
%}
%\ds{This point comess seems vague to me. Do we really want to include it? I'm happy with saying it's an open question whether the equivalence holds for larger/deeper networks, but not that happy to speculate why. The theory (SGLD) predicts it \emph{should} hold, right? If we want to say it doesn't shouldn't we point to some gap in that theory, as opposed to another theory?}

% While DIPs are asymptotically GPs, the SGD optimization with the DIP may be still preferable. This is because GP inference is expensive for high-resolution images. The memory usage is $O(n^2)$ and running time is $O(n^3)$ for exact inference where $n$ is the number of pixels (e.g., a 500$\times$500 image requires 233 GB memory). The DIP's memory requirements on the other hand scales linearly with the number of pixels, and inference with SGD is practical and efficient. This emphasizes the importance of the second part of this work which addresses several drawbacks of the vanilla SGD inference procedure making the DIP more robust and effective. % This does lead to a better algorithm for denoising and motivates the connection between the two parts [R2].

%% file: conclusion.tex
\section{Conclusion}
\label{conclusion}
\vspace{-2pt}

We presented a novel Bayesian view of the deep image prior, which parameterizes a natural image as the output of a convolutional network with random parameters and a random input. First,
%to understand the nature of such a prior, 
we showed that the output of a random convolutional network converges to a stationary zero-mean GP as the number of channels in each layer goes to infinity, and showed how to calculate the realized covariance. This characterized the deep image prior as approximately a stationary GP. Our work differs from prior work relating GPs and neural networks by analyzing the \emph{spatial} covariance of network activations on a single input image.
%which is exactly the structure we seek to understand for the deep image prior. 
We then use SGLD to conduct fully Bayesian posterior inference in the deep image prior, which improves performance and prevents the need for early stopping. Future work can further investigate the types of kernel implied by convolutional networks to better understand the deep image prior and the inductive bias of deep convolutional networks in learning applications.

\eat{
1. As the depth increases the covariance matrix becomes flatter, i.e. $K_d/K_0 \rightarrow 1, \forall d$. 
2. With random parameters the kernel size has no effect on the structure of the covariance function. Particularly for the iid input case $K_i = K_j, \forall i, j \neq 0$. (This seems odd but is true!)
3. The covariance matrix of the downsampled signal can be obtained by downsampling the covariance matrix of the original signal. Simlarly the covariance matrix of the upsampled signal can be obtained by upsampling the covariance matrix of the orignal signal. This way we can analyze the structure of the random convnets with pooling and unpooling layers (e.g., auto-encoder, UNet, ResNet). Note that unpooling introduces spatial correlations so kernel structure becomes more interesting in the deconvolution steps even for the iid input case.
4. The analysis can be easily extended to the 2D case.
}

%% file: acknowledgement.tex
\vspace{-10pt}
\paragraph{Acknowledgement} This research was supported in part
by NSF grants \#1749833, \#1749854, and \#1661259, and the MassTech Collaborative
for funding the UMass GPU cluster.

%% file: appendix.tex
\section*{\newaxv{Appendix}} \label{sec:appendix}

\def\swidthseven{0.140\linewidth}

%We complete the proof of Theorem 1 and show additional visualizations of the denoising and inpainting results.
%Our proof relies on the multivariate delta method, which we state for completeness below.

We provide the proof of Theorem 1 and show additional visualizations of the denoising and inpainting results.

\newtheoremstyle{named}{}{}{\itshape}{}{\bfseries}{}{.5em}{\thmnote{#3's}#1}
\theoremstyle{named}
\newtheorem*{namedtheorem}{Theorem}

\begin{proof}
Let $r = t_2 - t_1$. First, observe that
\begin{equation*}
\begin{split}
\bar{x}(t_1)^T \bar{x}(t_2)= \sum_{k=1}^n \bigg( \sum_{i=0}^{d-1}x(k, t_1-i)x(k, t_2-i) \bigg).\\
\end{split}
\end{equation*}
where $n$ is the number of channels in the input, $d$ is the filter width. 
By stationarity, the terms $x(k, t_1-i)x(k, t_2-i)$ are identically distributed with expected value $K_x(r)$, even though they are not necessarily independent. 
%Therefore $\E\left[\bar{x}(t_1)^T \bar{x}(t_2)\right] = dcK_x(t_1, t_2)$. 
Since the channels are drawn independently, the parenthesized expressions in the sum over $k$ are iid with mean $dK_x(r)$. 
% Therefore, the CLT applies and $\frac{1}{\sqrt{n}}\big(\bar{x}(t_1)^T \bar{x}(t_2) - d K_x(r)\big)$ converges in distribution to a Gaussian as $n \to \infty$. Thus 
According to Equation \ref{eq:2} in the main text, we have
$$
\begin{aligned}
K_z^{\rm erf} (t_1, t_2) &= \E_{x} V_{\rm erf}\big(\bar{x}(t_1), \bar{x}(t_2)\big) \\
&= \E_{x} \frac{2}{\pi} \sin^{-1} \frac{\sigma^2_u \bar{x}(t_1)^T\bar{x}(t_2)}{\sigma^2_u \|\bar{x}(t_1)\| \|\bar{x}(t_2)\|}. \\
\end{aligned}
$$

% Since $\bar{x}(t_1)^T\bar{x}(t_2)$ is asymptotically Gaussian one can show that 
% \begin{align}
% K_z^{\rm erf} (t_1, t_2) = \frac{2}{\pi}\sin^{-1} \frac{K_x(r)}{K_x(0)}, 
% \end{align}
% by applying the delta method~\cite{ver2012invented} 
% by applying the continuous mapping theorem. Thus the stationarity of $X$ implies that the limiting distribution of $Z$ is a stationary GP.

Consider the sequence $Y_n = (a_n, b_n, c_n)$ 

%\begin{equation}
%a_n = \frac{1}{n}\left(\bar{x}(t_1)^T\bar{x}(t_2)\right), \quad
%b_n = \frac{1}{n}\left(\bar{x}(t_1)^T\bar{x}(t_1)\right), \quad
%c_n = \frac{1}{n}\left(\bar{x}(t_2)^T\bar{x}(t_2)\right).
%\end{equation}

\begin{equation*}
\begin{split}
a_n &= \frac{1}{n}\left(\bar{x}(t_1)^T\bar{x}(t_2)\right) \\
b_n &= \frac{1}{n}\left(\bar{x}(t_1)^T\bar{x}(t_1)\right) \\
c_n &= \frac{1}{n}\left(\bar{x}(t_2)^T\bar{x}(t_2)\right),
\end{split}
\end{equation*}

%
% \begin{equation*}
% \begin{split}
% \bar{x}(t_1)^T \bar{x}(t_2) &= \sum_{k=1}^n \bigg( \sum_{i=0}^{d-1}x(k, t_1-i)x(k, t_2-i) \bigg).\\
% \end{split}
% \end{equation*}

%where $n$ is the number of channels in the network. 
By the strong law of large numbers we know the sequences $a_n$,  $b_n$ and $c_n$ each converge almost surely to their expected values $dK_x(t_1,t_2)$, $dK_x(t_1,t_1)$ and $dK_x(t_2,t_2)$, respectively. 
Since $K_x$ is stationary this is equal to $dK_x(r)$, $dK_x(0)$, and $dK_x(0)$, respectively, where $d$ is the filter width. Thus we have that $Y_n$ converges almost surely to $\bar{y} = \big(dK_x(r), dK_x(0), dK_x(0)\big)$.
% The covariance $\Sigma$ can be derived but is not important to compute the limiting distribution of the sequence $g(Y_n)$.

Consider the continuous function $g(y) = \frac{2}{\pi}\sin^{-1}\left( \frac{a}{\sqrt{bc}} \right)$ for $y = (a, b, c)$. Applying the continuous mapping theorem, we have that 

\begin{equation*}
    y_n \xrightarrow{a.s.} \mu \; \implies \; g(y_n) \xrightarrow{a.s.}  g(\mu)
\end{equation*}

Thus,

$$
K_z^{\rm erf} (t_1, t_2)  \xrightarrow{a.s.} \frac{2}{\pi}\sin^{-1} \frac{K_x(r)}{K_x(0)}.
$$

\end{proof}

%\dsin{Nit-picky question: if $a=0$ or $t^2 = 1$ then some entry of $\nabla g$ is zero? Do we need to worry about this case?}

\begin{figure*}
    \setlength{\tabcolsep}{1pt}
    \centering
    \begin{tabular}{ccccccc}
     (a) GT  & (b) Noisy GT & (c) SGD & (d) +Avg & (e) +Input & (f) +Input+Avg & (g) SGLD \\
    \includegraphics[width=\swidthseven]{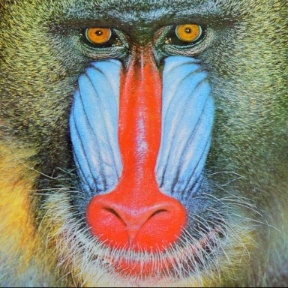} &
     \includegraphics[width=\swidthseven]{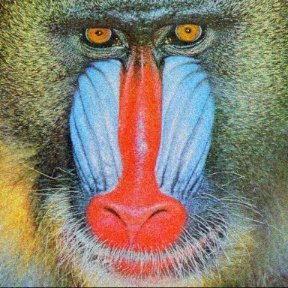} &
     \includegraphics[width=\swidthseven]{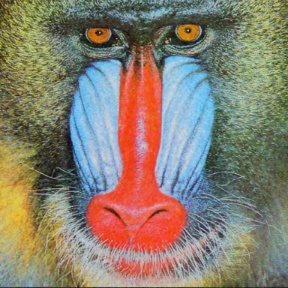} &
     \includegraphics[width=\swidthseven]{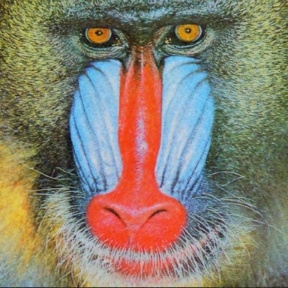} & 
     \includegraphics[width=\swidthseven]{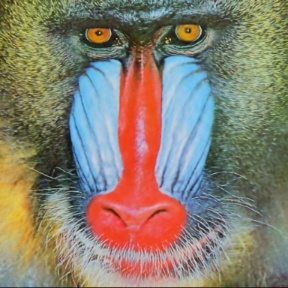} &
     \includegraphics[width=\swidthseven]{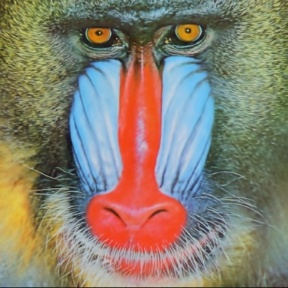} &
     \includegraphics[width=\swidthseven]{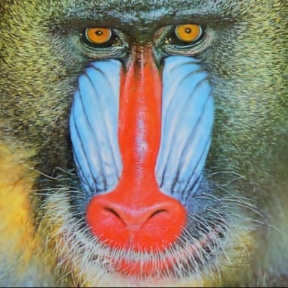} \\
     & 20.34 & 23.81 & 23.94 & 22.52 & 23.52 & \textbf{24.59} \\
         \includegraphics[width=\swidthseven]{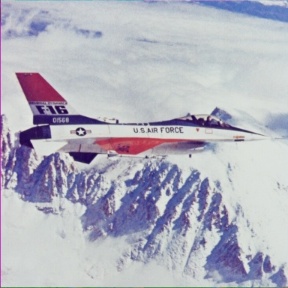} &
     \includegraphics[width=\swidthseven]{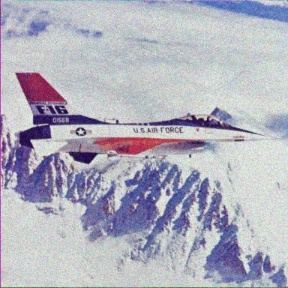} &
     \includegraphics[width=\swidthseven]{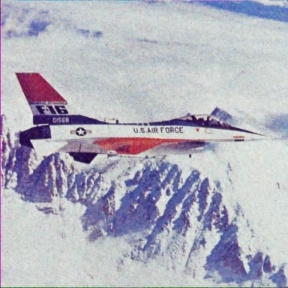} &
     \includegraphics[width=\swidthseven]{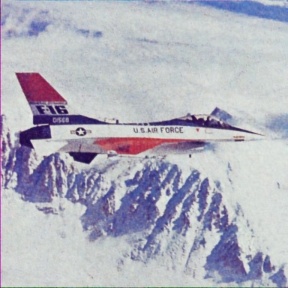} & 
     \includegraphics[width=\swidthseven]{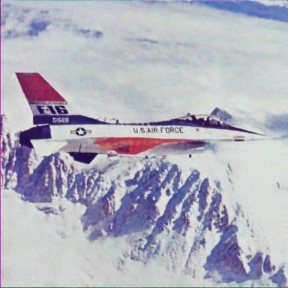} &
     \includegraphics[width=\swidthseven]{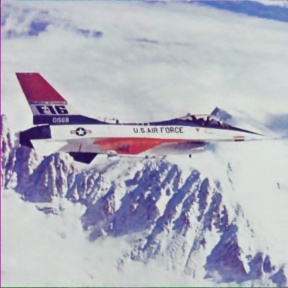} &
     \includegraphics[width=\swidthseven]{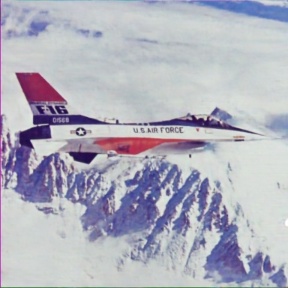} \\
     & 20.33 & 29.81 & 31.12 & 30.65 & 32.73 &\textbf{ 32.95 }\\
         \includegraphics[width=\swidthseven]{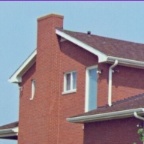} &
     \includegraphics[width=\swidthseven]{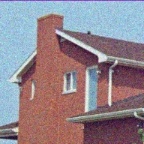} &
     \includegraphics[width=\swidthseven]{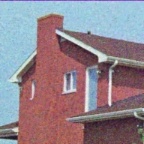} &
     \includegraphics[width=\swidthseven]{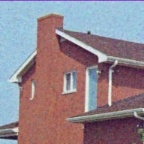} & 
     \includegraphics[width=\swidthseven]{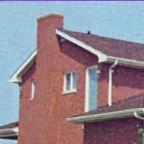} &
     \includegraphics[width=\swidthseven]{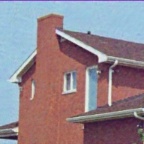} &
     \includegraphics[width=\swidthseven]{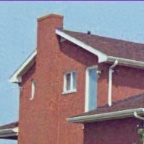} \\
     & 20.28 & 26.14 & 28.59 & 27.92 & 30.37 & \textbf{31.32} \\
         \includegraphics[width=\swidthseven]{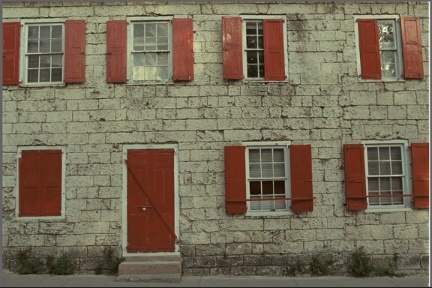} &
     \includegraphics[width=\swidthseven]{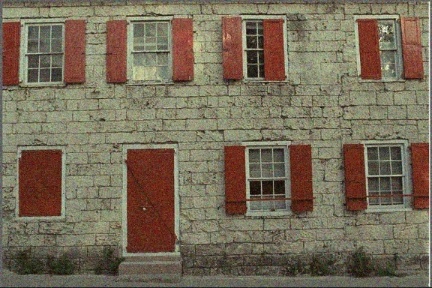} &
     \includegraphics[width=\swidthseven]{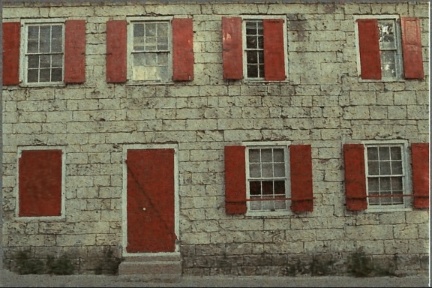} &
     \includegraphics[width=\swidthseven]{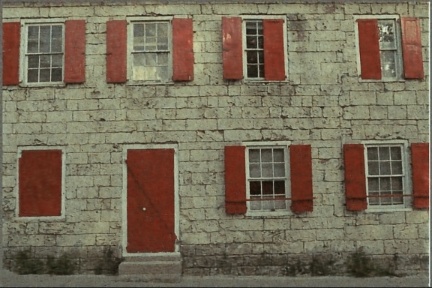} & 
     \includegraphics[width=\swidthseven]{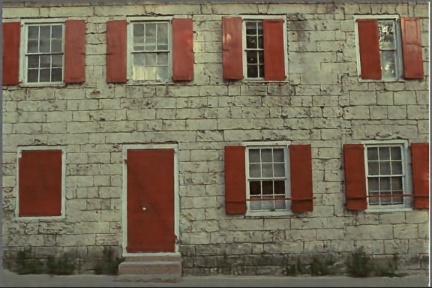} &
     \includegraphics[width=\swidthseven]{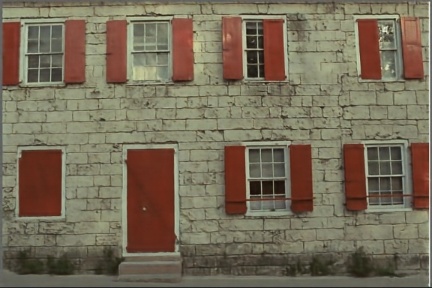} &
     \includegraphics[width=\swidthseven]{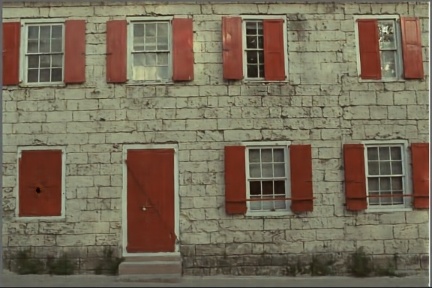} \\
     & 20.28 & 26.67 & 27.16 & 26.28 & 27.25 & \textbf{27.95} \\
         \includegraphics[width=\swidthseven]{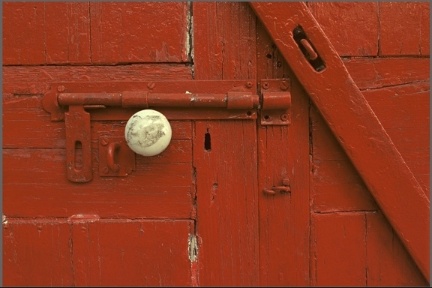} &
     \includegraphics[width=\swidthseven]{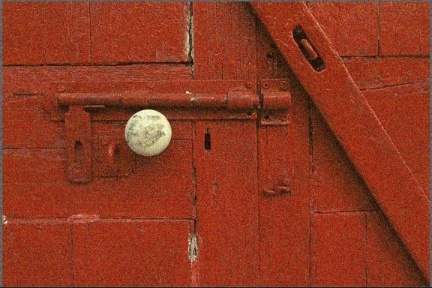} &
     \includegraphics[width=\swidthseven]{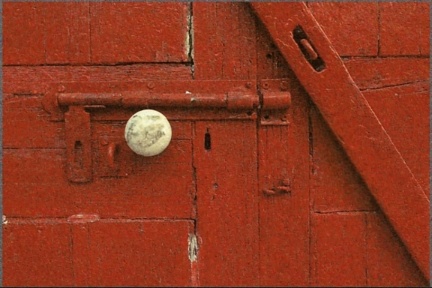} &
     \includegraphics[width=\swidthseven]{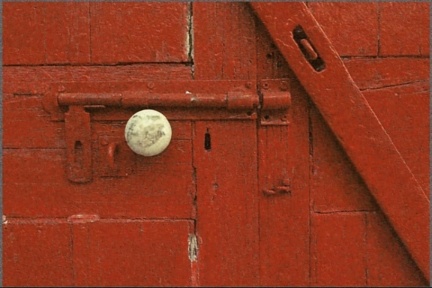} & 
     \includegraphics[width=\swidthseven]{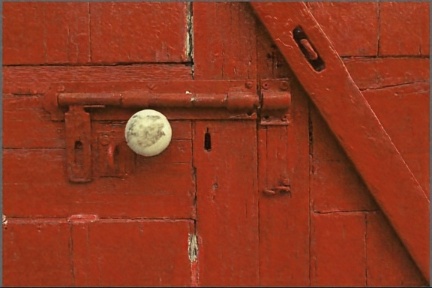} &
     \includegraphics[width=\swidthseven]{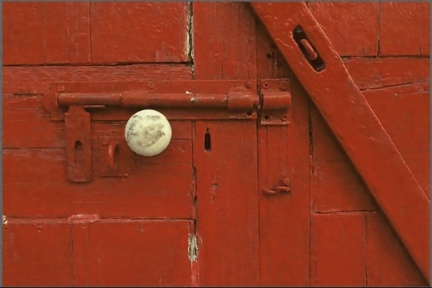} &
     \includegraphics[width=\swidthseven]{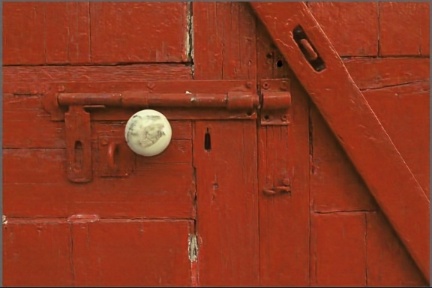} \\
     & 20.57 & 28.77 & 29.87 & 30.30 & 31.74 & \textbf{32.05} \\
         \includegraphics[width=\swidthseven]{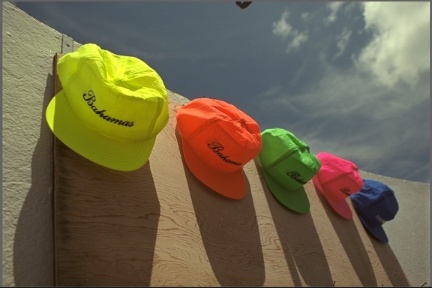} &
     \includegraphics[width=\swidthseven]{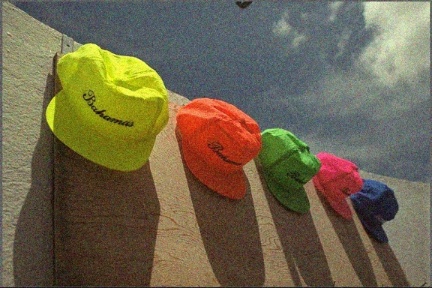} &
     \includegraphics[width=\swidthseven]{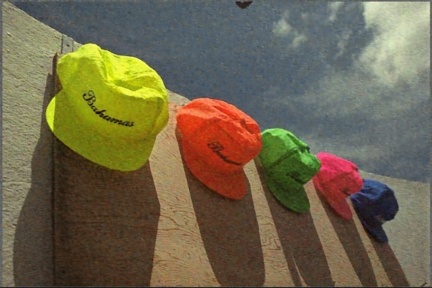} &
     \includegraphics[width=\swidthseven]{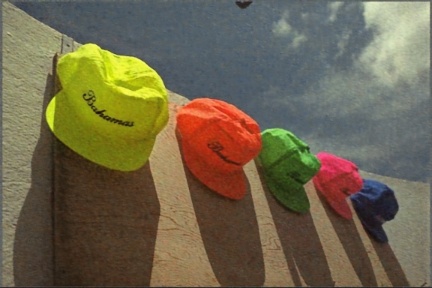} & 
     \includegraphics[width=\swidthseven]{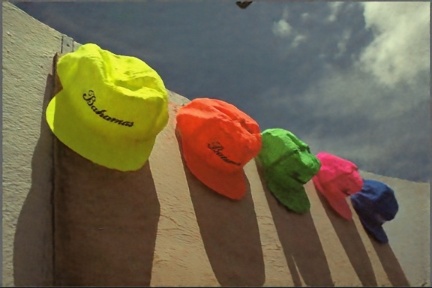} &
     \includegraphics[width=\swidthseven]{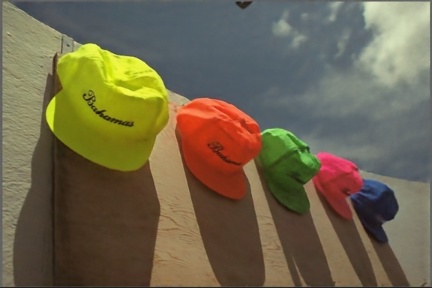} &
     \includegraphics[width=\swidthseven]{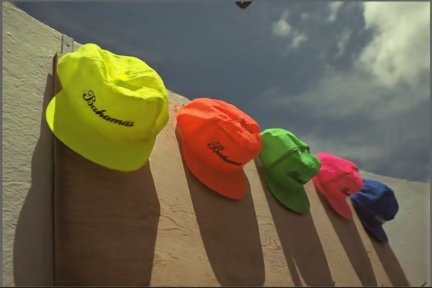} \\
     & 20.36 & 30.23 & 31.02 & 31.34 & 32.84 & \textbf{33.39} \\
         \includegraphics[width=\swidthseven]{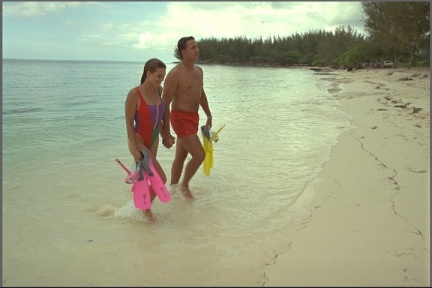} &
     \includegraphics[width=\swidthseven]{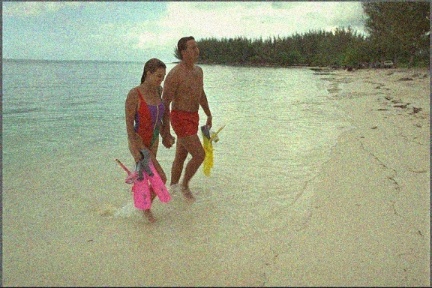} &
     \includegraphics[width=\swidthseven]{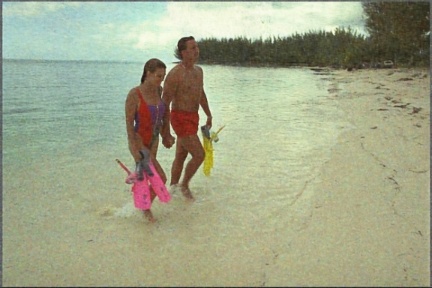} &
     \includegraphics[width=\swidthseven]{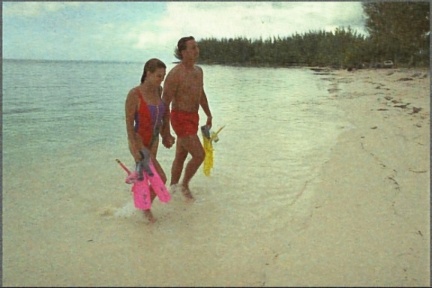} & 
     \includegraphics[width=\swidthseven]{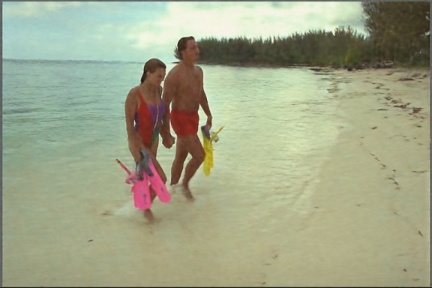} &
     \includegraphics[width=\swidthseven]{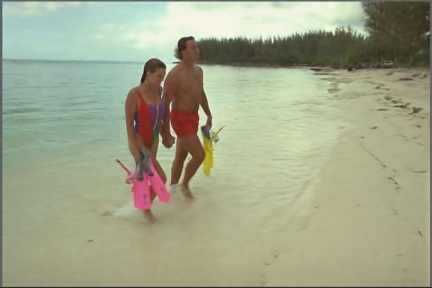} &
     \includegraphics[width=\swidthseven]{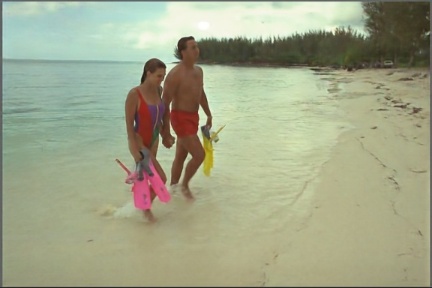} \\
     & 20.32 & 29.62 & 30.46 & 30.84 & 32.07 & \textbf{32.81} \\
         \includegraphics[width=\swidthseven]{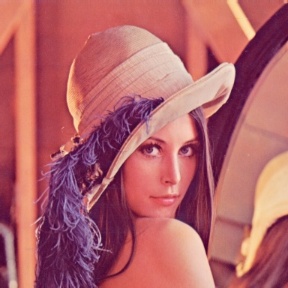} &
     \includegraphics[width=\swidthseven]{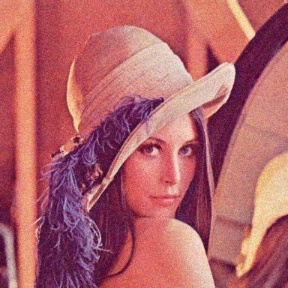} &
     \includegraphics[width=\swidthseven]{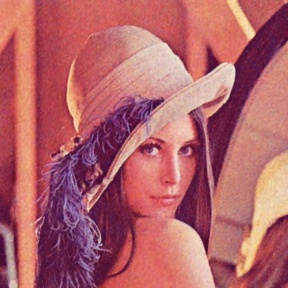} &
     \includegraphics[width=\swidthseven]{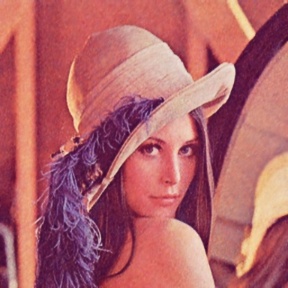} & 
     \includegraphics[width=\swidthseven]{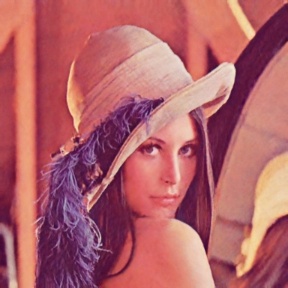} &
     \includegraphics[width=\swidthseven]{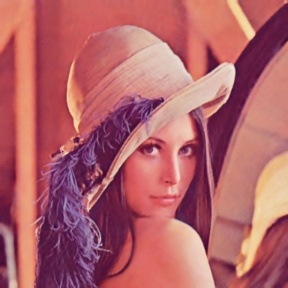} &
     \includegraphics[width=\swidthseven]{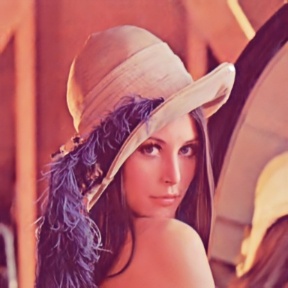} \\
     & 20.37 & 29.07 & 30.28 & 30.21 & 31.81 & \textbf{32.08} \\ 
    \end{tabular}
    \caption{\emph{(Best viewed magnified.)} \textbf{Comparison of inference schemes for image denoising.} From left to right each column shows \textbf{(a)} ground truth image, \textbf{(b)} the ground truth image corrupted with Gaussian noise of $\sigma=25$, \textbf{(c)} inference with SGD and early stopping (iteration picked at maximum PSNR), \textbf{(d)} previous step with exponential window average, \textbf{(e)} SGD with input noise and early stopping, \textbf{(f)} the previous procedure with exponential window averaging, and \textbf{(g)} SGLD. Underneath each figure the PSNR values are listed. Note that except SGLD every procedure requires early stopping.
    \label{fig:denoising}
    }
\end{figure*}

\begin{figure*}
    \setlength{\tabcolsep}{1pt}
    \centering
    \begin{tabular}{ccccccc}
      (a) Corrupted GT & (b) SGD & (c) +Avg. & (d) +Input & (e) +Input+Avg & (f) SGLD  & (g) Uncertainty\\
    \includegraphics[width=\swidthseven]{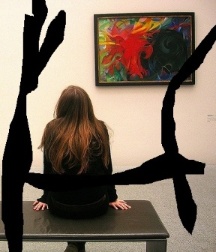} &
    \includegraphics[width=\swidthseven]{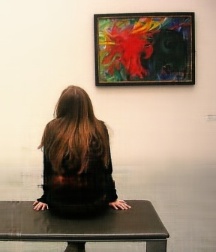} &
    \includegraphics[width=\swidthseven]{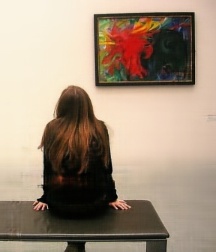} &
    \includegraphics[width=\swidthseven]{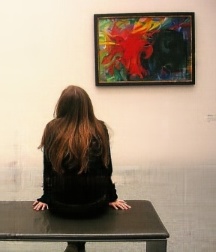} &
    \includegraphics[width=\swidthseven]{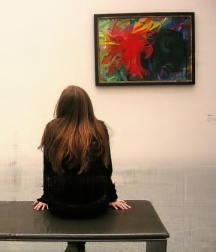} &
    \includegraphics[width=\swidthseven]{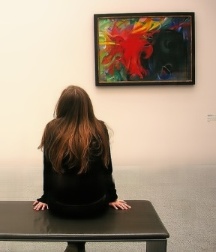} &
    \includegraphics[width=\swidthseven]{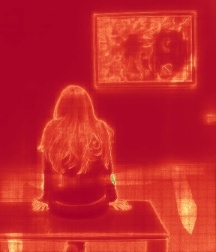} \\
     &  \small{27.86} &  \small{28.06} &  \small{28.51} &  \small{28.53} &  \small{\textbf{31.43}} &\\ 
    \includegraphics[width=\swidthseven]{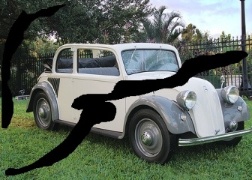} &
    \includegraphics[width=\swidthseven]{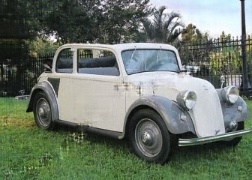} &
    \includegraphics[width=\swidthseven]{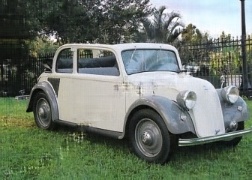} &
    \includegraphics[width=\swidthseven]{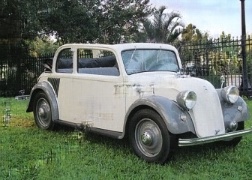} &
    \includegraphics[width=\swidthseven]{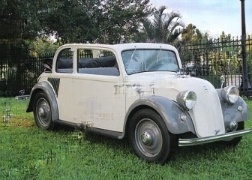} &
    \includegraphics[width=\swidthseven]{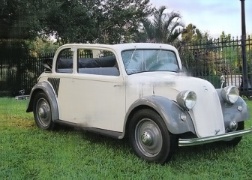} &
    \includegraphics[width=\swidthseven]{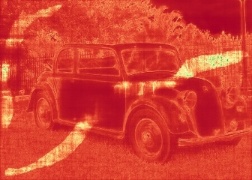} \\
     &  \small{22.85} &  \small{22.97} &  \small{23.31} &  \small{23.57} &  \small{\textbf{27.10}} &\\ 
         \includegraphics[width=\swidthseven]{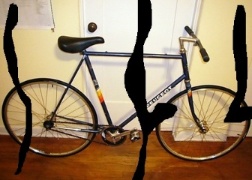} &
    \includegraphics[width=\swidthseven]{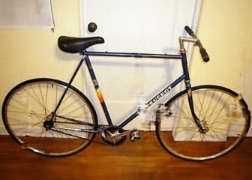} &
    \includegraphics[width=\swidthseven]{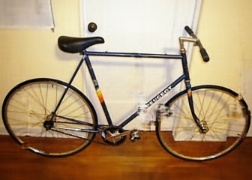} &
    \includegraphics[width=\swidthseven]{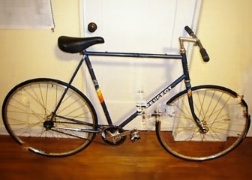} &
    \includegraphics[width=\swidthseven]{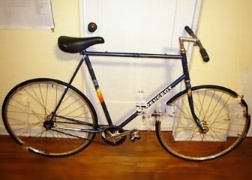} &
    \includegraphics[width=\swidthseven]{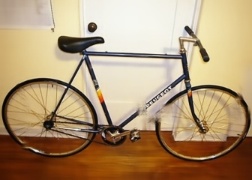} &
    \includegraphics[width=\swidthseven]{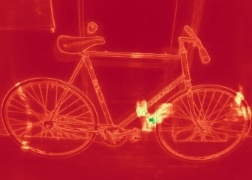} \\
     &  \small{25.35} &  \small{26.52} &  \small{26.48} &  \small{26.74} &  \small{\textbf{28.87}} &\\ 
         \includegraphics[width=\swidthseven]{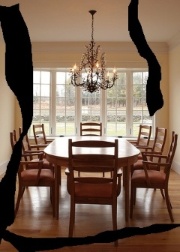} &
    \includegraphics[width=\swidthseven]{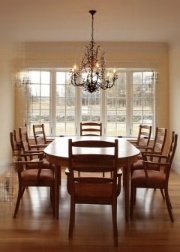} &
    \includegraphics[width=\swidthseven]{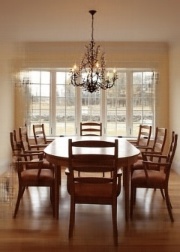} &
    \includegraphics[width=\swidthseven]{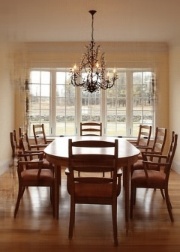} &
    \includegraphics[width=\swidthseven]{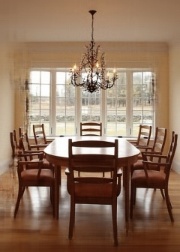} &
    \includegraphics[width=\swidthseven]{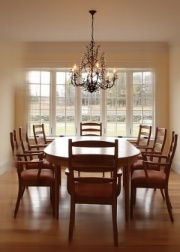} &
    \includegraphics[width=\swidthseven]{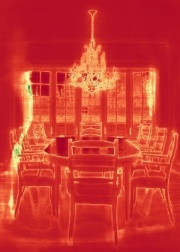} \\
     &  \small{28.16} &  \small{28.18} &  \small{28.71} &  \small{29.04} &  \small{\textbf{30.78}} &\\ 
         \includegraphics[width=\swidthseven]{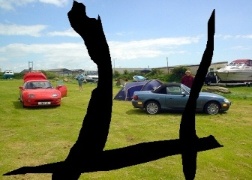} &
    \includegraphics[width=\swidthseven]{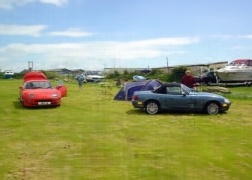} &
    \includegraphics[width=\swidthseven]{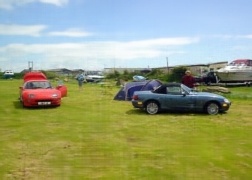} &
    \includegraphics[width=\swidthseven]{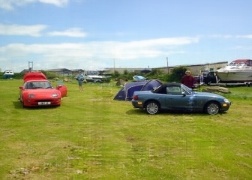} &
    \includegraphics[width=\swidthseven]{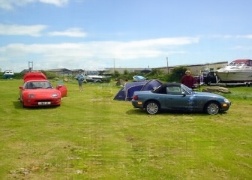} &
    \includegraphics[width=\swidthseven]{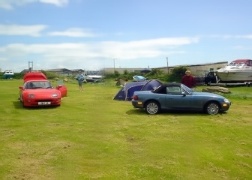} &
    \includegraphics[width=\swidthseven]{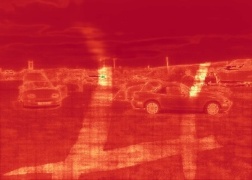} \\
     &  \small{26.22} &  \small{26.29} &  \small{26.47} &  \small{26.66} &  \small{\textbf{29.05}} &\\ 
         \includegraphics[width=\swidthseven]{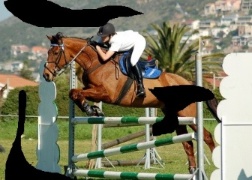} &
    \includegraphics[width=\swidthseven]{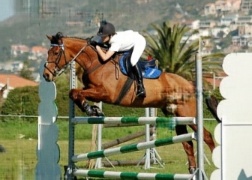} &
    \includegraphics[width=\swidthseven]{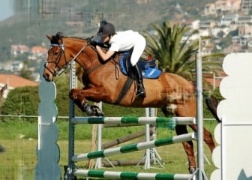} &
    \includegraphics[width=\swidthseven]{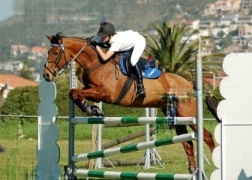} &
    \includegraphics[width=\swidthseven]{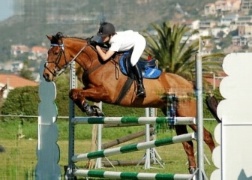} &
    \includegraphics[width=\swidthseven]{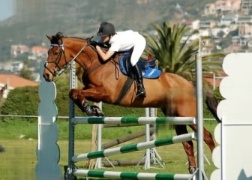} &
    \includegraphics[width=\swidthseven]{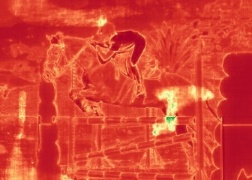} \\
     &  \small{26.04} &  \small{26.09} &  \small{26.11} &  \small{26.28} &  \small{\textbf{28.41}} &\\ 
         \includegraphics[width=\swidthseven]{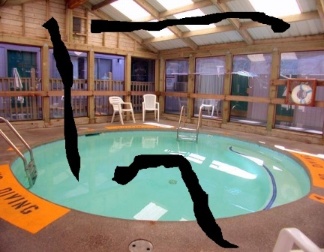} &
    \includegraphics[width=\swidthseven]{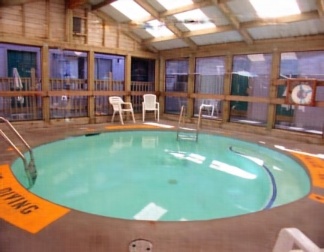} &
    \includegraphics[width=\swidthseven]{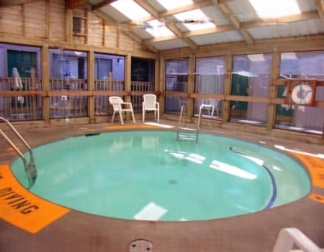} &
    \includegraphics[width=\swidthseven]{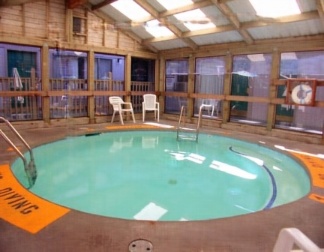} &
    \includegraphics[width=\swidthseven]{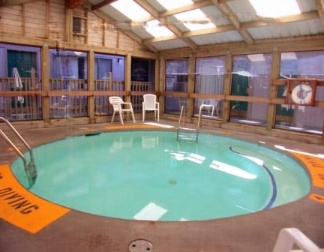} &
    \includegraphics[width=\swidthseven]{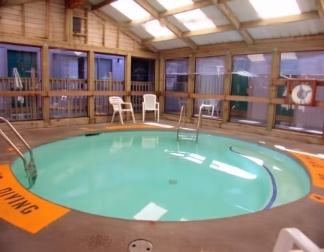} &
    \includegraphics[width=\swidthseven]{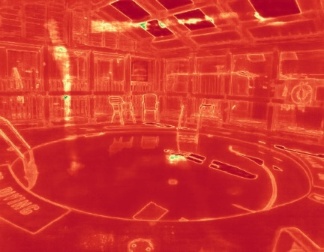} \\
     &  \small{29.03} &  \small{29.15} &  \small{30.15} &  \small{30.33} &  \small{\textbf{32.95}} &\\ 
    \end{tabular}
     \caption{\emph{(Best viewed magnified.)} \textbf{Comparison of inference schemes for image inpainting.} \textbf{(a)} The input image with parts masked out. \textbf{(b-f)} Results using various inference schemes (see the previous figure and the main text for details of the methods). \textbf{(f)} SGLD achieves higher PSNR (shown underneath each figure) than the baseline methods \textbf{(b-e)}. The early stopping iteration is set as 2000 for the first two schemes \textbf{(b, c)} and 3000 for the third and fourth methods \textbf{(d, e)}. For these experiments we use a 6-layer auto-encoder without skip connections and random noise as input as reported in~\cite{ulyanov18deep}. \textbf{(g)} The variance estimated from the posterior samples visualized as a heat map. Notice that the uncertainty is low in the missing regions that are surrounded by areas of relatively uniform appearance such as wall, sky and swimming pool, and higher in non-uniform areas such as those near the boundaries of different object in the image.}
    \label{fig:inpainting}
\end{figure*}

\paragraph{Additional visualizations.} Figure \ref{fig:denoising} and \ref{fig:inpainting} present results for various inputs using various inference schemes for image denoising and image inpainting respectively. 

